\documentclass[12pt]{article}

\usepackage{amsmath}
\usepackage{amssymb}
\usepackage{bbm}
\usepackage{color}
\usepackage{url} 
\usepackage{mathtools}
\usepackage{times}
\usepackage{bm}
\usepackage{natbib}
\usepackage[plain,noend]{algorithm2e}

\DeclarePairedDelimiter\floor{\lfloor}{\rfloor}

\newtheorem{theorem}{Theorem}[section]
\newtheorem{lemma}[theorem]{Lemma}

\newtheorem{condition}{Condition}

\newenvironment{proof}{\paragraph{Proof:}}{\hfill$\square$}

\begin{document}

{
  \title{\bf \bf Random Planted Forest: A Directly Interpretable Tree Ensemble}
  \author{Munir Hiabu \hspace{.2cm}\\
    Department of Mathematical Sciences, University of Copenhagen\\
    and \\
    Enno Mammen\thanks{
    We gratefully acknowledge support by the Deutsche Forschungsgemeinschaft (DFG) through the Research Training Group RTG 1953.} \\
    Institute for Applied Mathematics, Heidelberg University\\
    and \\
    Joseph T. Meyer\footnotemark[1] \\
    Institute for Applied Mathematics, Heidelberg University}
  \maketitle
}

\bigskip
\begin{abstract}
    We introduce a novel interpretable tree based algorithm for prediction in a regression setting. Our motivation is to estimate the unknown regression function from a functional decomposition perspective in which the functional components correspond to lower order interaction terms. The idea is to modify the random forest algorithm by keeping certain leaves after they are split instead of deleting them. This leads to non-binary trees which we refer to as planted trees. An extension to a forest leads to our random planted forest algorithm. Additionally, the maximum number of covariates which can interact within a leaf can be bounded. If we set this interaction bound to one, the resulting estimator is a sum of one-dimensional functions. In the other extreme case, if we do not set a limit, the resulting estimator and corresponding model place no restrictions on the form of the regression function.
	In a simulation study we find encouraging prediction and visualisation properties of our random planted forest method.
    We also develop theory for an idealized version of random planted forests in cases where the interaction bound is low. We show that if it is smaller than three, the idealized version achieves asymptotically optimal convergence rates up to a logarithmic factor. Code is available on GitHub \url{https://github.com/PlantedML/randomPlantedForest}.
\end{abstract}

\noindent%
\vfill

\newpage


\section{Introduction}
	In many ways machine learning has been very disruptive in the last two decades. The class of neural networks has shown unprecedented and previously unthinkable predictive accuracy in fields such as image recognition \citep{rawat2017deep,lecun2015deep}, speech recognition \citep{hinton2012deep}, and natural language processing \citep{collobert2011natural}.
	Deep neural networks are strong in applications where huge amounts of data can be assembled and many variables interact with each other. A second disruptive class of machine learning algorithms are decision tree ensembles.
	The gradient boosting machine \citep{friedman2001greedy} in particular excels in many applications.
	It often is the best performing algorithm for tabular data \citep{grinsztajn2022tree} and as such it is praised by many practitioners. Among the 29 challenge-winning solutions posted on Kaggle during 2015, 17 used xgboost -- a variant of a gradient boosting machine \citep{chen2016xgboost}. The second most popular method, deep neural networks, was used in 11 solutions. A downside of such algorithms is that they are black box methods in the sense that they are hard to visualize and interpret on the one hand, and lack theoretical backing on the other.
	
	These black box methods are in contrast to classical statistical models that assume an explicit structure such as a linear \citep{nelder1972generalized} or additive model \citep{friedman1981projection,buja1989linear}. 
	Estimators received from classical statistical models are highly accurate if the model is correctly specified. However, they perform poorly if the data deviates strongly from the structure assumed by the model.
	Our aim is to combine the best of the two worlds. In a regression set-up we consider a regression problem and assume that the regression function can be approximated well by a function satisfying the bounded interaction BI($r$) condition for a predefined small $r\in\mathbb{N}$.
	
\begin{condition}[BI($r$)] A function $m:\mathbb{R}^d\rightarrow\mathbb{R}$ satisfies BI($r$) if it can be written as
	\begin{align}
	    m(x)=\sum_{t\in T_r
	    }m_t(x_t)\label{anova1}
	\end{align}
	for functions $m_t:\mathbb{R}^{|t|}\rightarrow\mathbb{R}$, where $x_t=(x_k)_{k \in t}$ and $T_r=\{t\subseteq\{1,\dots,d\}\ |\ |t|\leq r\}$.
	\end{condition}
	
	Equivalently, we write $m(x)=\sum_{t\in T_r}m_t(x)$ for functions $m_t:\mathbb{R}^{d}\rightarrow\mathbb{R}$, where $m_t$ only depends on the values of $x_t$, $t\subseteq\{1,\dots,d\}$. We will use both versions interchangeably throughout this work. For a function that satisfies BI($r$), the components $m_t$ are not identified without additional constraints. For our work, the constraints used are of secondary importance. Possible constraints are deferred to the supplementary material. Note that $m_\emptyset$ is a constant. Thus if $r<<d$ we assume that the regression function can be well approximated by a functional decomposition including only low dimensional structures \citep{stone1994use,hooker2007generalized,chastaing2012generalized}. Observe that if $r\geq d$, we impose no additional restraint to the function $m$. We propose a new tree based algorithm we call random planted forests (rpf) which given any $r$ satisfies BI($r$).
 
\begin{figure}
		\includegraphics[scale=0.8]{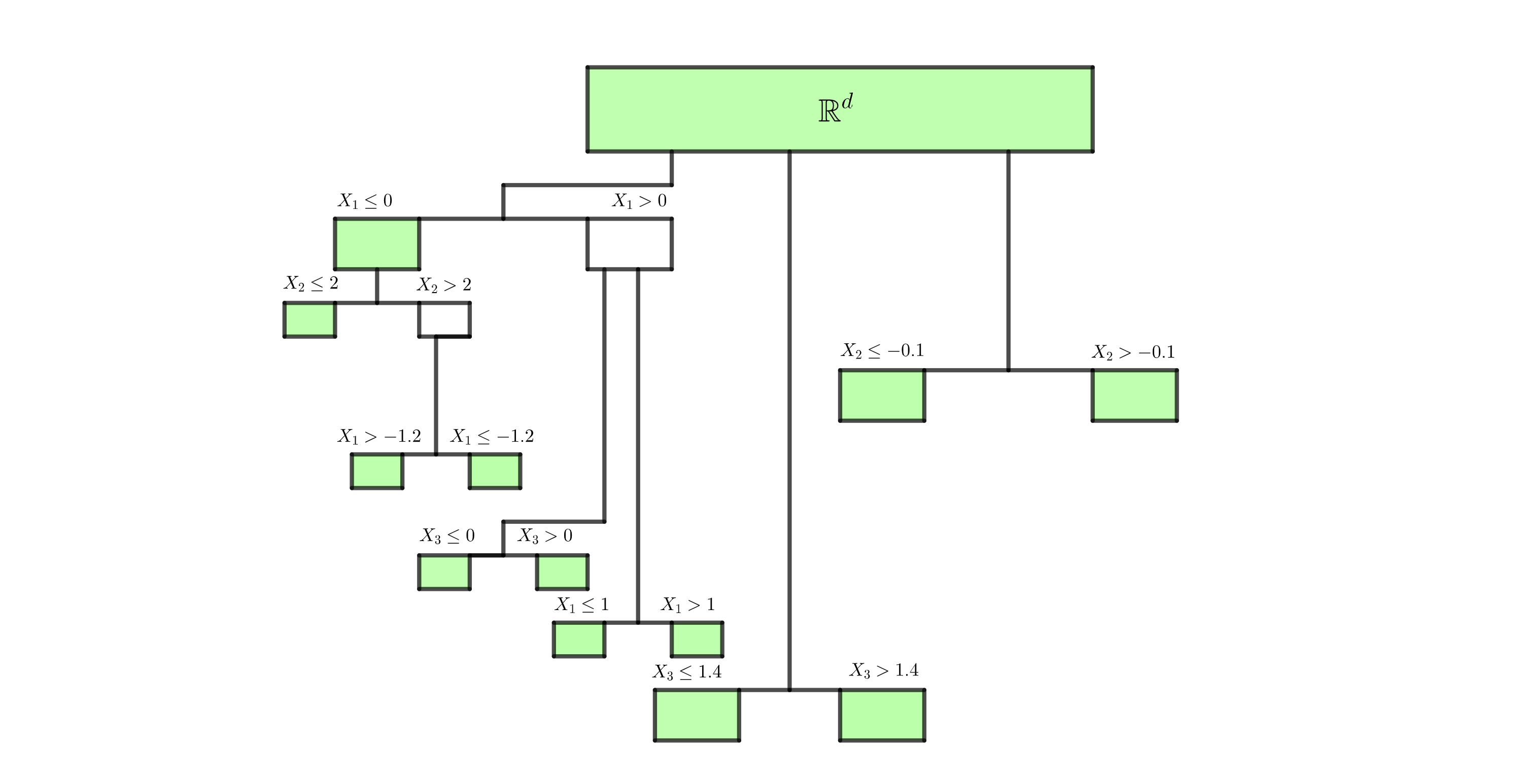}
		\renewcommand{\thefigure}{1a}
		\caption{Illustration of a planted tree. The height of boxes indicate the order in which splits occurred. Green boxes correspond to leaves, which are still splittable in the future. When splitting a leaf $\mathcal{I}\subseteq\mathbb{R}^d$ into two new leaves $\mathcal{I}=\mathcal{I}^+\cup\mathcal{I}^-$ with respect to a coordinate which was not used in the construction of $\mathcal{I}$, the leaf $\mathcal{I}$ turns to a white box which represents an inner node, which can not be used for further splitting. In this example, the resulting estimator depends on 3 components $x_1,x_2,x_3$ and satisfies BI(2).}
		\label{ill:interaction}
	\end{figure}

	A tree in rpf can be thought of as a single traditional decision tree in which a leaf is not removed from the algorithm when split in some cases, see Rule (a) in Subsection \ref{Setup}. Furthermore, if a leaf was created using $\texttt{max\_interaction}$ coordinates, we only allow for splits with respect to coordinates which were already used, see Rule (b) in Section \ref{Setup}. Here $\texttt{max\_interaction}\in\mathbb{N}$ is a tuning parameter. These rules are the main ingredients of rpf and they lead to a non-binary tree structure. We also made use of further rules that have been proposed in other implementations of random forests, see Rules (a)—(f) in Section \ref{Setup} for a full description of rpf. The resulting estimator satisfies BI($\texttt{max\_interaction}$). Figure \ref{ill:interaction} provides a short illustration.
    A simple heuristic why it may be beneficial not to remove a leaf from a tree when splitting is the following. Consider the regression problem $Y_i=m(X_i)+\epsilon_i$ with $i=1,\dots,n$ and $m(x)=\sum_{k=1}^d\mathbbm{1}(x_k\leq 0.5)$ for large $d$.
	If we never remove the original leaf, one can approximate $m$ by splitting the original leaf once with respect to each covariate $k$. Thus for each covariate $n$ data points are considered for finding the optimal split value.  Additionally we end up with on average  $n/2$ data points in each leaf.
    In the original random forests algorithm, in order to find a similar function one would have to grow a tree where each leaf is constructed by splitting once with respect to each covariate. This implies that we end up with $2^d$ leaves, which on average contain $n/(2^d)$ data points. Thus the estimation should be much worse both because of less precise split points and less accurate fit values.
    
    This example shows that rpf behaves quite differently than classical random forests independent of the interaction bound $\texttt{max\_interaction}$ imposed on the algorithm. In the simulation study presented in Section \ref{simulations}, we find that even in the case where no interaction bound is set ($\texttt{max\_interaction}\geq d$) the results are promising.

\begin{figure}
		\includegraphics[scale=0.6]{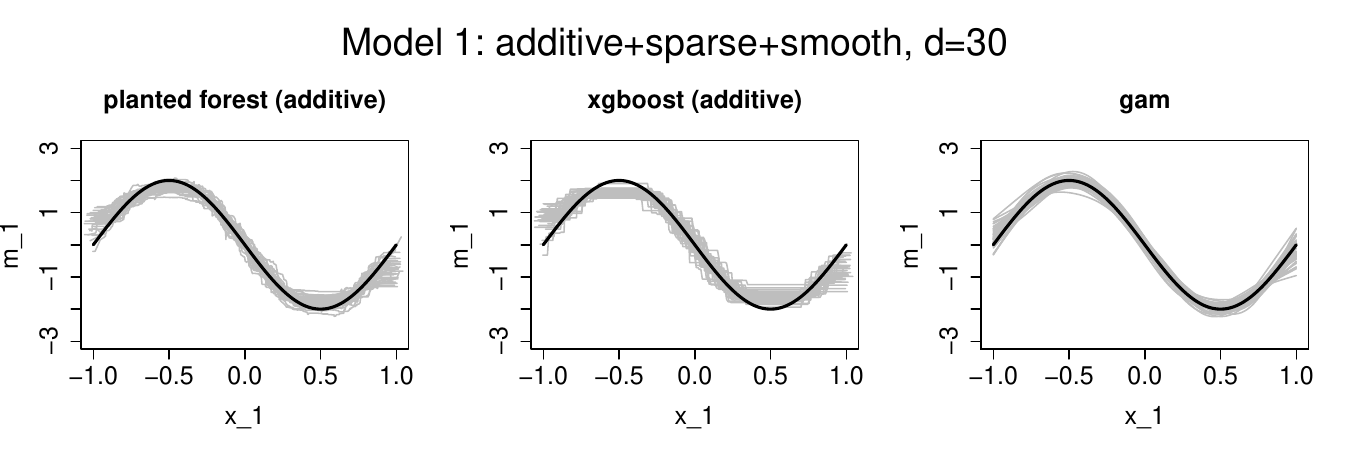}
		\caption{Estimates from 40 simulations for $m_1$ of Model 1: $m(x_1,\dots,x_{30})=m_1(x_1)+m_2(x_2)$  with $m_k(x_k)=2(-1)^k\sin(\pi x_k)$. The true function is visualized as a black solid line. Sample size is $n=500$. Predictors have an approximate pairwise correlation of $0.3$ and the noise has variance $1$. For rpf and xgboost, parameters are picked from a grid search. The gam curves have data-driven parameters. } \label{fig:addsmooth}
	\end{figure}

\begin{figure}
		\includegraphics[scale=0.6]{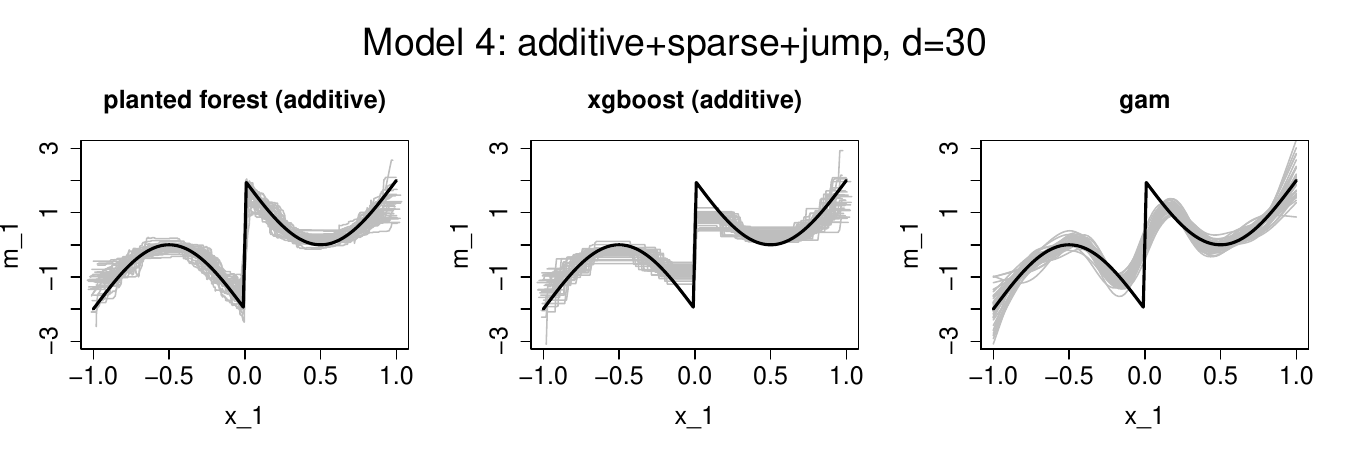}
		\caption{Estimates from 40 simulations for $m_1$ of Model 4: $m(x_1,\dots,x_{30})=m_1(x_1)+m(x_2)$ with $m_k(x_k)=( 2(-1)^k\sin(\pi x_k)-2)\mathbbm1 (x\geq0) + (-2\sin(\pi x_k)+2)\mathbbm1 (x<0) $. The true function is visualized as a black solid line. Sample size is $n=500$. Predictors have an approximate pairwise correlation of $0.3$ and the noise has variance $1$.  For rpf and xgboost parameters are picked from a grid search. The gam curves have data-driven parameters. }
		\label{fig:addjump}
	\end{figure}

\begin{figure}
		\includegraphics[scale=0.6]{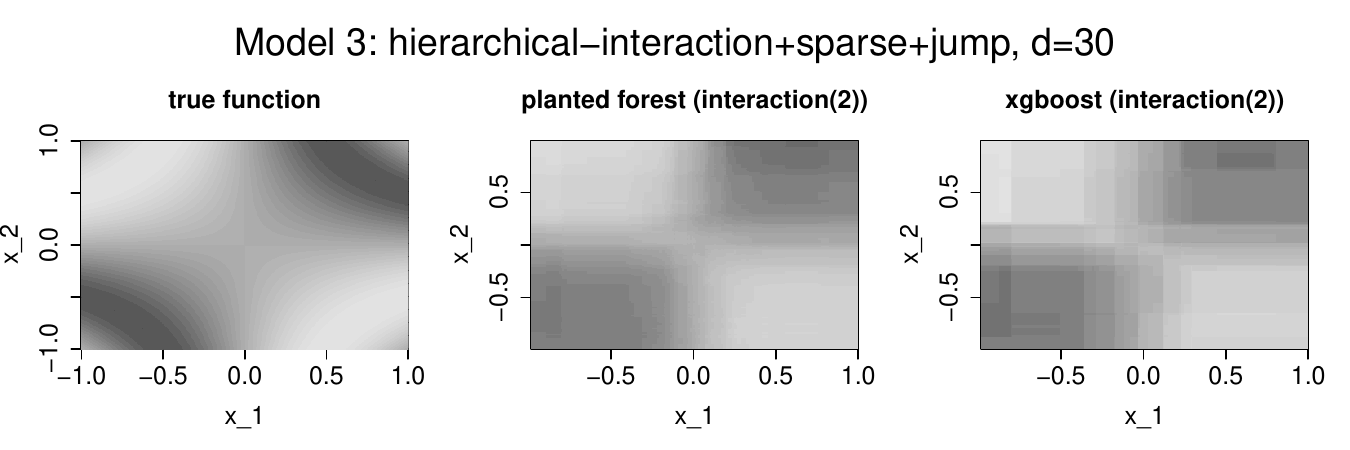}
		\caption{Heatmap from the median performing run for each method, measured via mean squared error, out of 40 simulations. Predictions are for $m_{1,2}$ of Model 3: $m(x_1,\dots,x_{30})=\sum_{k=1}^3m_k(x)+\sum_{1\leq k<j\leq 3}m_{k,j}(x_k,x_j)$ with $m_{k,j}(x_k,x_j)= 2(-1)^k\sin(\pi x_kx_j)$, $m_k(x_k)=(-1)^k 2\sin(\pi x_1)$. Sample size is $n=500$. The predictors have an approximate pairwise correlation of $0.3$ and the noise has variance $1$. Parameters are picked from a grid search.}
		\label{fig:h}
	\end{figure}

    In practice, the value $r$ of the underlying regression function is unknown. 
    Bounding the order of interaction $\texttt{max\_interaction}<<d$ does increase the performance of rpf if the true model satisfies BI$(r)$, $r\leq\texttt{max\_interaction}$. 
    An additional advantage of choosing $\texttt{max\_interaction}=1$ or $\texttt{max\_interaction}=2$ is that predictions can be visualized easily. Figures \ref{fig:addsmooth} -- \ref{fig:h} show plots that visualize one and two dimensional components of a high dimensional model respectively. The models we use for comparison were chosen since they are the strongest competitors in the simulation results in Section \ref{simulations} which satisfy BI($1$) or BI($2$). 
	
	In the past, other algorithms satisfying BI($r$) for certain $r=1,\dots,d$ have been developed. Rpf differs from the tree based explainable boosting machine \citep{lou2012intelligible,lou2013accurate,caruana2015intelligible,lengerich2020purifying} and the neural network based  neural additive model \citep{agarwal2020neural} recently introduced in several ways. Similar to rpf, these methods aim to approximate the regression function via the functional decomposition (\ref{anova1}) for $r=1$ or $r=2$. However, they rather resemble classical statistical methods described earlier. After specifying a fixed structure, the explainable boosting machine assumes that every component is relevant and all components are fit via a backfitting algorithm \citep{breiman1985estimating}. A similar principle using backpropagation instead of backfitting is applied in the neural additive model.
    In contrast, the rpf algorithm may ignore certain components completely. Additionally, when considering $r=2$ interaction terms do not need to be specified beforehand.
	Hence explainable boosting machine and the neural additive model do not share two key strengths of rpf: a) automatic interaction detection up to the selected order, b) strong performance in sparse settings, see Section \ref{simulations}. Some model-based boosting algorithms \citep{hofner2014model} have similar properties to the ones given above. In particular, they also suffer from the fact that they do not have automatic interaction detection, which is especially troublesome if a large number of covariates are present. An algorithm not suffering from this problem is multivariate adaptive regression splines \citep{friedman1991multivariate}.  However, it has a continuous output. Thus it does not adapt well when single jumps are present. Furthermore, as seen in our simulation study in Section \ref{simulations}, multivariate adaptive regression splines are less accurate when many covariates are active. Considering the random forests algorithm, \cite{tan2022cautionary} find that it performs poorly in models which satisfy BI(1). They propose a modification of random forests where each tree is replaced by a finite fixed number of trees that grow in parallel. For a discussion of the performance of rpf in such models, see also the theory developed in an earlier version of this paper Hiabu, Mammen and Meyer (2020).
 
	
    From a theoretical point of view, the most comparable algorithm is random forests. A theoretical study of Breiman's original version of random forests \citep{breiman2001random} is rather complex due to the double use of data, once in a CART splitting criterion and once in the final fits. \cite{scornet2015consistency} provide a consistency proof for Breiman's algorithm while assuming an BI(1). They also show that the forest adapts to sparsity patterns. In more recent papers, theory has been developed for subsampled random forests. By linking to theory of infinite order U-statistics, asymptotic unbiasedness and asymptotic normality has been established for these modifications of random forests, see \cite{mentch2016quantifying, mentch2017formal}, \cite{wager2018estimation}, \cite{peng2019asymptotic}. In our mathematical analysis of rpf we assume that the splitting values do not strongly depend on the response variables which excludes using the CART splitting criterion. In this respect we follow a strand of literature on random forests, see e.g. \cite{biau2008consistency} and \cite{biau2012analysis}. Thus we circumvent theoretical problems caused by the double use of data as in Breiman's random forest. We leave it to future research to study the extent to which theory introduced here carries over to random planted forests with data-dependent splitting rules, in particular for the case of random planted forests based on subsampling or sample-splitting.
    
    In summary, rpf is useful for the following reasons: Rpf is a tree based algorithm that given an $r\in\mathbb{N}$ satisfies BI($r$). In contrast to most tree based algorithms, this enables interpretability of the regression fucntion by choosing a small $r$. Another tree based algorithm that can be set to satsify BI($r$) for any $r$ is gradient boosting machine by setting the tree depth to $r$. However, our simulation study shows that small $r$ in xgboost does not always lead to satisfactory results. In general, intuitively, the difference between gradient boosting and rpf is that gradient boosting iteratively updates an estimator globally, while rpf approximates the components locally. Thus we conjecture that rpf performs better in the presense of highly irregular functional components, while gradient boosting may have an edge in highly regular settings. In this sense, we believe rpf to be complementary to existing algorithms. Our simulation study shows a good overall performance of rpf. The theoretical analysis suggests almost optimal convergence rates for rpf in low interaction settings. The theory developed is interesting in itself since it shows optimal convergence rates for tree based algorithms satisfying the conditions given in Section \ref{Theoretical properties in the additive case}.
	
\section{Random Planted Forests}

\subsection{Notation on Trees}
A tree produces an estimator of the form $\sum_{\mathcal{I}} m_\mathcal{I}\mathbbm{1}(x\in \mathcal{I})$ for some sets $\mathcal{I}\subseteq\mathbb{R}^d$ which we refer to as \textit{leaves}. The set $\mathcal{I}=\mathbb{R}^d$ is referred to as \textit{root}. We \textit{split} leaves by setting $\mathcal{I}^+:=\{x\in\mathcal{I}\ | x_k>c\}$ and $\mathcal{I}^-:=\{x\in\mathcal{I}\ |\ x_k\leq c\}$ for some $k=1,\dots,d$, $c\in\mathbb{R}$. These new leaves are either included in the estimator or replace $\mathcal{I}$. A set $\mathcal{I}$ which was replaced is no longer referred to as leaf but as \textit{inner node}. 
\subsection{Setup}\label{Setup}

\begin{figure}
		\includegraphics[scale=0.8]{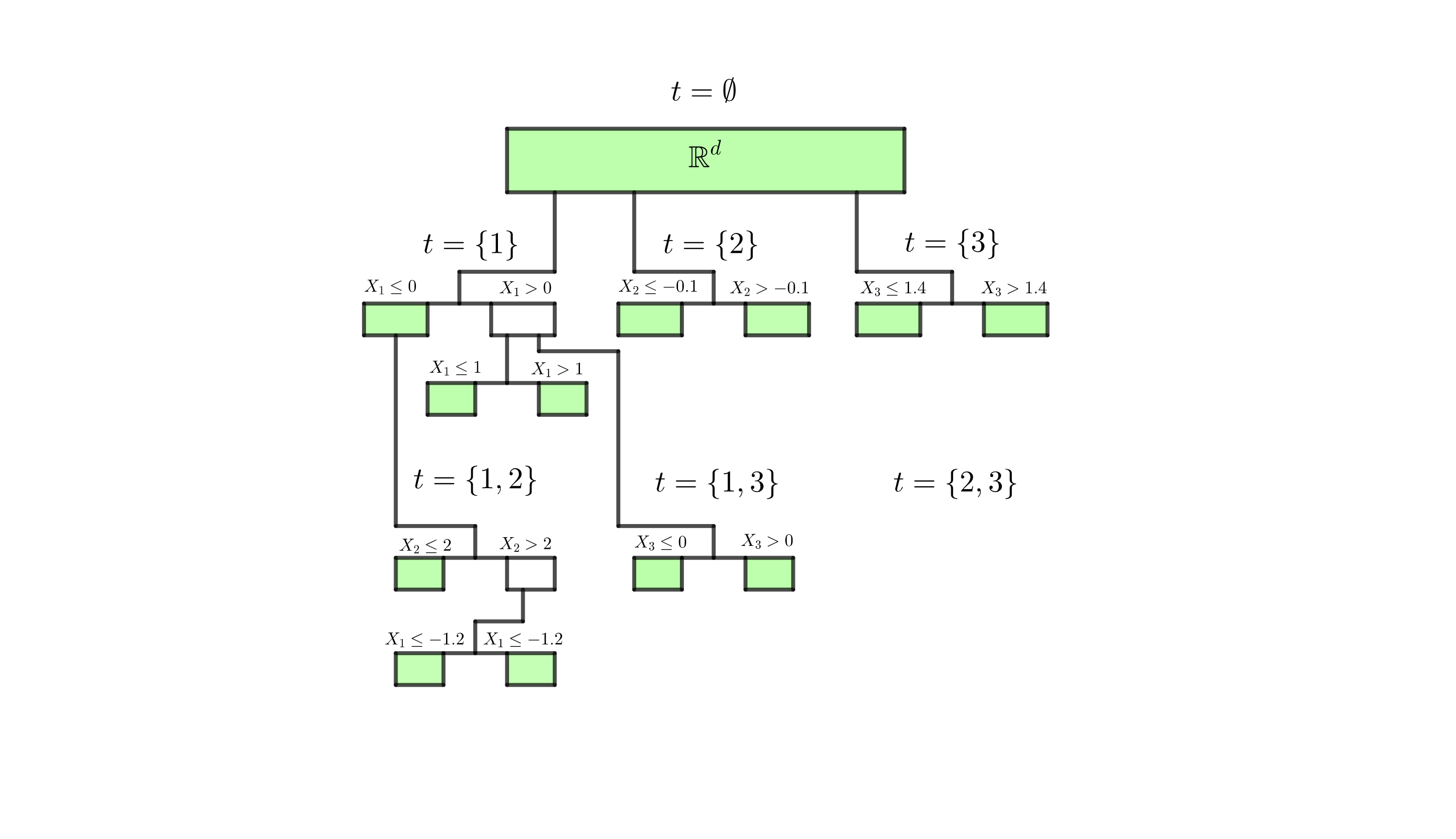}
		\renewcommand{\thefigure}{1b}
		\caption{ Alternative illustration of the planted tree from Figure \ref{ill:interaction}. Nodes and leaves are sorted into leaf types $t$. Note that the height of boxes no longer indicates the order in which splits occurred.}
		\label{ill:leafType}
	\end{figure}

We are handed data $(Y_i, X_{i,1},\dots,X_{i,d})_{i=1}^n$ consisting of i.i.d. observations with $Y_i,X_{i,k}\in\mathbb{R}$ and consider the regression problem 
	\[
	    E[Y_i|X_i=x] = m(x),
	\]
	with the goal of estimating $m$. Growing a planted tree in rpf is similar to the construction of a tree in random forests. It depends on a parameter $r=\texttt{max\_interaction}$, which corresponds to the parameter $r$ in BI($r$). In order to obtain a decomposition of the form \eqref{anova1}, we sort leaves into \textit{leaf types}, depending on which variables $t\in T_r$ were used to construct the respective leaf. Rpf differs from random forests in the following way.
	\begin{itemize}
	    \item[(a)] When splitting a leaf $\mathcal{I}\subseteq\mathbb{R}^d$ into two new leaves $\mathcal{I}=\mathcal{I}^+\cup\mathcal{I}^-$ with respect to a coordinate which was not used in the construction of $\mathcal{I}$, the leaf $\mathcal{I}$ is not deleted and may be used for splitting again in the future.
	    \item[(b)] Leaves with leaf type $|t|=r$ may only be split with respect to dimensions $k\in t$.
	    \item[(c)] If a leaf is updated, the corresponding estimator is updated by adding the average residual within the leaf.
	\end{itemize}
	For example, the root will never be deleted during a planted tree construction. An illustration of the algorithm unfolding is given in Figure \ref{ill:interaction}. Since a leaf $\mathcal{I}$ may be split multiple times, the resulting final leaves of a planted tree are typically not disjoint. Thus simply defining the estimator as the sample average within a leaf does not yield useful results. Instead we keep track of the estimator during the algorithm and update it by averaging residuals in each iteration step similar to gradient boosting or backfitting. Since the order in which the splits occur makes a difference, we indicate the order by vertical position in Figure \ref{ill:interaction}. Note that (a) is important for obtaining a useful estimator satisfying BI($r$). For an intuition, assume (b) for $r=1$ and leaves are deleted after splitting as is the case for random forests. Then each tree will only depend on 1 variable, which does not result in a useful estimator. As discussed above, leaves are sorted into leaf types $t\in T$. An alternative illustration of rpf highlighting this sorting is given in Figure \ref{ill:leafType}.
	Additionally, the following methods are implemented which are typical for forest type algorithms. 
	\begin{itemize}
	    \item[(d)] A forest is obtained by averaging the estimators from trees grown from $\texttt{ntrees}\in\mathbb{N}$ bootstrap samples.
	    \item[(e)] Dimensions $k$ and leaves $\mathcal{I}$ used for splitting are selected from a random subset of possible dimension-leaf combinations. The parameter controlling the size of the subset is $\texttt{t\_try}\in(0,1]$.
	    \item[(f)] Split values are chosen from a set of $\texttt{split\_try}\in\mathbb{N}$ random points.
	\end{itemize}
	Thus, the algorithm depends on five tuning parameters: The number of splits $\texttt{nsplits}\in\mathbb{R}$, the number of trees $\texttt{ntrees}\in\mathbb{N}$, the order of interaction $\texttt{max\_interaction}\in\mathbb{N}$ and the parameters $\texttt{t\_try}\in(0,1]$, $\texttt{split\_try}\in\mathbb{N}$. The importance of the parameters is discussed in the supplement.
	
    We proceed by explaining the algorithm rigorously. A planted tree consists of a finite set $\{(\mathcal{I}_{t,l},m_{t,l})\ |\ t\in T_r,\ l=1,\dots,L_t\}$ of leaves $\mathcal{I}_{t,l}$ and corresponding values $m_{t,l}\in\mathbb{R}$. Observe that we sort the leaves with respect to their leaf type $t\in T_r$. Here $L_t$ denotes the number of leaves with type $t$. For $t\in T_r$ and $l=1,\dots,L_t$ we have
	\begin{align}
	    \mathcal{I}_{t,l}=\prod_{k=1}^dA_{t,l,k}, \label{equation hyperrectangle}
	\end{align}
	where  $A_{t,l,k} \in \{A\subseteq\mathbb{R}\ |\ A=(b,c] \ \text{or}\ A=(b,\infty)\ \text{for}\ b\in[-\infty,\infty), c\in(b,\infty)\}$ for $k \in t$  and  $A_{t,l,k}=\mathbb{R}$ for $k\notin t$. 
	Define the estimator for $m$ by $\widehat{m}(x)=\sum_{t\in T_r}\widehat{m}_t(x)$ with
	\[
	    \widehat{m}_t(x)=\sum_{l=1}^{L_t}m_{t,l}\mathbbm{1}(x\in \mathcal{I}_{t,l}).
	\]
	Note that for any $x\in\mathbb{R}^d$, due to \eqref{equation hyperrectangle}, $\widehat{m}_t$ only depends on the values of $x_t$. Thus $\widehat{m}$ satisfies BI($r$). Next, we introduce the calculation of a split in Subsection \ref{Calculating a split} followed by the iteration procedure in Subsection \ref{Planted tree }. We conclude with the extension to a forest and a discussion on the driving parameters.
	
\subsection{Calculating a Split}\label{Calculating a split}
	
	We are handed a leaf $\mathcal{I}_{t,l}$ of the form \eqref{equation hyperrectangle}, a value $m^*$, a splitting coordinate $k\in\{1,\dots,d\}$, a splitting value $c\in\mathbb{R}$, and residuals $R_1,\dots,R_n$. Define a partition $\mathcal{I}_{t,l}=\mathcal{I}^+\cup\mathcal{I}^-$ by $\mathcal{I}^+:=\{x\in\mathcal{I}_{t,l}\ | x_k>c\}$ and $\mathcal{I}^-:=\{x\in\mathcal{I}_{t,l}\ |\ x_k\leq c\}.$
	Let $g_{+}:=\sum_{X_i\in \mathcal I^+}R_i/\sum_{X_i\in \mathcal I^+}1$, $g_{-}:=\sum_{X_i\in \mathcal I^-} R_i/\sum_{X_i\in \mathcal I^-}1$ and
	\[
	    g:\mathbb{R}^d\rightarrow\mathbb{R},\ g(x)= \mathbbm{1}(x\in \mathcal I^+)g_{+}+\mathbbm{1}(x\in \mathcal I^-)g_{-}.
	\]
	Observe that $g$ is constant on each of the sets $\mathcal{I}^+$, $\mathcal{I}^-$, and $\mathbb{R}^d\backslash\mathcal{I}_{t,l}$. We update the values values $m_{+} :=m^*+g_{+}$, $m_{-} :=m^*+g_{-}$ and residuals
	$
	R_i^{\text{new}}:=R_i-g(X_i)
	$.
	Pseudo-code of this procedure is given in Algorithm \ref{Algorithm: RPF 1}.

    \begin{algorithm}[H]     
    \SetAlgoLined
	    \caption{Calculating a Split}
  \SetKwData{Left}{left}\SetKwData{This}{this}\SetKwData{Up}{up}
\SetKwFunction{Union}{Union}\SetKwFunction{FindCompress}{FindCompress}
\SetKwInOut{Input}{Input}\SetKwInOut{Output}{Output}
			  \textbf{Input} {$\mathcal I_{t,l}, m^*,k,c, R_1,\dots,R_n, X_{1},\dots,X_{n}$}\\
		\textbf{Calculate} $\mathcal{I}^+,\mathcal{I}^-$
			\For{$i=1,\ldots,n$} {
		           $R_i:=R_i-g(X_i)$}
	\textbf{Calculate}  $ m_{+}:= m^*+g_{+}$; $\quad m_{-}:=m^*+g_{-}$ \\
	  \textbf{Output} {$ m_{+}$, $m_{-}$, $\mathcal{I}^+$, $\mathcal{I}^-$, $R_1,\dots,R_n$}
   \label{Algorithm: RPF 1}
    \end{algorithm}
 
\subsection{Random Planted Tree}\label{Planted tree }

	In addition to $r=\texttt{max\_interaction}$, the algorithm depends on $\texttt{nsplits}\in\mathbb{N}$ which determines the number of iterations before the algorithm stops. We start off by setting the residuals $R_i^{0}:=Y_i$ for $i=1,\dots, n$. We define the root $I^{0}_{\emptyset,1}:=\mathbb{R}^d$ with value $m^{0}_{\emptyset,1}(x)=0$ and thus the number of leaves with type $t=\emptyset$ is $L_\emptyset^{0}=1$. Since we start off with no other leaves, for $t\neq \emptyset$ we have $L_{t}^{0}=0$.
 
	Each iteration step is carried out as follows. In step $s\geq 1$ we are handed leaves with values $\{(\mathcal{I}^{s-1}_{t,l},m^{s-1}_{t,l})\ |\ t\in T_r,\ l=1,\dots,L^{s-1}_t\}$ and residuals $R_1^{s-1},\dots,R_n^{s-1}$.	
	The leaves are of the form \eqref{equation hyperrectangle}. Assume for now that we have selected a leaf $\mathcal{I}^{s-1}_{t_s,l_s}$, a coordinate $k_s=1,\dots, d$, a split point $c_s\in\mathbb{R}$ and use Algorithm \ref{Algorithm: RPF 1}. The chosen combination must be viable in the sense that $k_s\in t_s$ if $|t_s|=r$. Now, there are two cases with different updating procedures.
	\begin{itemize}
	    \item If $k_s\in t_s$, we use $m^*:=m_{t_s,l_s}^{s-1}$ as an input for Algorithm \ref{Algorithm: RPF 1}. Then the resulting leaves $\mathcal{I}^+$ and $\mathcal{I}^-$ are of type $t_s$. We replace $\mathcal{I}^{s-1}_{t_s,l_s}$ by $\mathcal{I}^+$ and $\mathcal{I}^-$ in the set of leaves of type  $t_s$.
	    \item If $k_s\notin t_s$, we use $m^*:=0$ as an input for Algorithm \ref{Algorithm: RPF 1}. The leaves $\mathcal{I}^+$ and $\mathcal{I}^-$ obtained are of type $t_s\cup \{k_s\}\neq t_s$. We keep $\mathcal{I}^{s-1}_{t_s,l_s}$ in the set of leaves of type  $t_s$ and add leaves $\mathcal{I}^+,\mathcal{I}^-$ to the set of leaves of type $t_s\cup \{k_s\}$.
	\end{itemize}  
	The values $m_+$ and $m_-$ are set to be the corresponding values of $\mathcal{I}^+$ and $\mathcal{I}^-$ respectively. The residuals are updated accordingly. All other leaves and values are taken over from step $s-1$. In order to select $\mathcal{I}^{s-1}_{t_s,l_s}$, $k_s$, and $c_s$ we use the CART methodology. Thus we set
	\begin{align}
	    (t_s,l_s,k_s,c_s):=\underset{t,l,k,c}{\arg\ \min}\sum_{i=1}^n(R^{t,l,k,c}_i)^2\label{equation2},
	\end{align} 
	where for $i=1,\dots,n$ we denote by $R^{t,l,k,c}_i$ the residual one obtains by using Algorithm \ref{Algorithm: RPF 1} with inputs $\mathcal{I}^{s-1}_{t,l}$, $k$, $c$, and $m^* := m^{s-1}_{t,l}$ if $k\in t$, 
	$m^* :=0$ otherwise. The algorithm stops after $\texttt{nsplits}$ iterations. Pseudo-code of this procedure is given in Algorithm \ref{Algorithm: RPF 2}. 
	
    \begin{algorithm}[H]  
    \SetAlgoLined 
		\caption{Growing a Random Planted Tree}
		 \textbf{Tuning parameters} $\texttt{max\_interaction}$; $\quad \texttt{nsplits}$\\
			 \textbf{Input} $(Y_1,X_1),\dots, (Y_n,X_n)$\\
			\For{$i=1,\dots,n$} {$R_i\leftarrow Y_i$}
		$\mathcal{I}_{\emptyset,1}\leftarrow\mathbb{R}^d$; $\quad m_{\emptyset,1}\leftarrow 0$; $\quad L_{\emptyset}\leftarrow 1$\\
			\lFor{$t\neq\emptyset$}  {$L_{t}\leftarrow 0$}
			\For{$s={1,\dots, \texttt{nsplits}}$}{
		\textbf{Calculate} $t_s,l_s,k_s, c_s$ using Equation (\ref{equation2})  \\
			\uIf{$k_s\in t_s$}{
			$m^*\leftarrow m_{t_s,l_s}$\;
			\textbf{Calculate} $m_{+},m_{-}$, $\mathcal{I}^+,\mathcal{I}^-$, $R_1,\dots, R_n$ using Algorithm \ref{Algorithm: RPF 1}\;
			 $\mathcal{I}_{t_s,l_s}\leftarrow \mathcal{I}^+$; $\quad \mathcal{I}_{t_s,L_{t_s}+1}\leftarrow \mathcal{I}^-$
			 $m_{t_s,l_s}\leftarrow m_{+}$; $\quad m_{t_s,L_{t_s}+1}\leftarrow m_{-}$; $\quad L_{t_s}\leftarrow L_{t_s}+1$
                }\Else{
			  $m^*\leftarrow 0$ \\
		      \textbf{Calculate} $m_{+},m_{-}$, $\mathcal{I}^+,\mathcal{I}^-$, $R_1,\dots, R_n$ using Algorithm \ref{Algorithm: RPF 1}
			$t\leftarrow t_s\cup k_s$
			 $\mathcal{I}_{t,L_{t}+1}\leftarrow \mathcal{I}^+$; $\quad \mathcal{I}_{t,L_{t}+2}\leftarrow \mathcal{I}^-$
		 $m_{t,L_{t}+1}\leftarrow m_{+}$; $\quad m_{t,L_{t}+2}\leftarrow m_{-}$; $\quad L_{t}\leftarrow L_{t}+2$
			}
			} 
		\textbf{Output} $\{(\mathcal{I}^{(s-1)}_{t,l},m^{(s-1)}_{t,l})\ |\ t\in T_r,\ l=1,\dots,L^{(s-1)}_t\}$ 
	\label{Algorithm: RPF 2}
	\end{algorithm}

\subsection{Random Planted Forests}\label{From a tree  to a Forest}

	The random planted forests algorithm depends on additional tuning parameters $\texttt{ntrees}$ which denotes the number of trees in the forest, $\texttt{t\_try}$ which corresponds to the $\texttt{m\_try}$ parameter from random forests, and $\texttt{split\_try}$ which imposes a mechanism similar to extremely random forests \citep{geurts2006extremely}.
	
	More precisely, in order to reduce variance, we extend Algorithm \ref{Algorithm: RPF 2} to a forest estimator similar to the random forest procedure. We draw $\texttt{ntrees}$ independent bootstrap samples $(Y_1^v,X_1^v), \dots, (Y_n^v,X_n^v)$ from our original data. On each of these bootstrap samples, we apply Algorithm \ref{Algorithm: RPF 2} with two minor adjustments. In each iteration step, given a coordinate $k$ and a leaf $\mathcal{I}_{t,l}$, we uniformly at random select $\texttt{split\_try}$ values $C_{t,l,k}$ (with replacement)  from  
	\[
	    \big\{X_{i,k}\ \big|\ X_{i}\in\mathcal{I}_{t,l},\ X_{i,k}\neq \max\{X_{i',k}\ |\ X_{i'}\in \mathcal{I}_{t,l}\}\big\}.
	\] 
	Secondly, define a set representing the viable combinations
	\[
	    B:=\big\{(t,k)\in T_r \times \{1,\dots,d\}\ \big|\ k\in t,\ \max\{L_t,L_{t\backslash k}\big\}\geq 1\}.
	\]
	In each iteration step we select a subset $M\subseteq B$ uniformly at random, where $|M|=\lceil|B|\cdot\texttt{t\_try}\rceil$ for some given value $\texttt{t\_try}\in(0,1]$.
	In step 4, when calculating $(t_s,l_s,k_s,c_s)$ in Equation (\ref{equation2}), we minimize under the additional conditions that $c\in C_{t,l,k}$ and $(t,k)\in M$ or $(t\backslash k,k)\in M$. The resulting estimators are denoted by $\widehat{m}_t^v(x) := \sum_{l=1}^{L^v_t}\mathbbm{1}(x\in\mathcal{I}^v_{t,l})m^v_{t,l}$ for $v=1,\dots, \texttt{ntrees}$. The forest estimator is given by $\widehat{m}(x)=\sum_{t\in T_r}\widehat{m}_t(x)$ with
	\[
	\widehat{m}_t(x):=\frac{1}{\texttt{ntrees}}\sum_{v=1}^\texttt{ntrees} \widehat{m}_t^v(x).
	\]
	Further remarks on the algorithm including discussions on the tuning parameters can be found in Section \ref{Section: The Random Planted Forest Algorithm} in the supplement.
\section{Simulations}\label{simulations}

	In this section we conduct an extensive simulation study in order to arrive at an understanding of how rpf copes with different settings and how it compares to other methods. For each of 100 Monte-Carlo simulations $s=1,\dots,100$ we consider the regression setup
	\[
	Y_i^s= m(X_i^s) +\varepsilon_i^s, \quad i=1,\dots, 500,
	\]
	where $\varepsilon^s_i \overset{\mathrm{i.i.d.}}{\sim} N(0,1)$.
	We consider twelve different models outlined in Table \ref{models}. Following \cite{nielsen2005smooth}, the predictors $(X^s_{i,1}, \dots, X^s_{i,d})$ are distributed as follows.
	We first generate $(\widetilde X^s_{i,1}, \dots, \widetilde X^s_{i,d})$ from a $d$-dimensional standard multi-normal distribution with mean equal to 0 and $\textrm{Cov}(\widetilde  X^s_{i,j},\widetilde  X^s_{i,k})=\textrm{Corr}(\widetilde  X^s_{i,j},\widetilde X^s_{i,k})=0.3$ for $j\neq k$. Then we set
	\[
	X^s_{i,k}=2.5\pi^{-1}\text{arctan}(\widetilde X^s_{i,k}).
	\]
	This procedure is repeated independently 500 times.
	\begin{table}[ht]
		\centering
		\caption{Dictionary for model descriptions. In total we consider 12 models. Each model has a model structure (first six rows) as well as a function shape (last two rows). The constant $d$ denotes the total number of predictors. In sparse models we use $d=4, 10, 30$. For dense models $d=4,10$ is considered.} \label{models}
		\begin{tabular}{|ll|}
			\hline
			Description & Meaning\\ 
			\hline
			\hline
			additive+sparse& 
			$m(x)=m_1(x_1)+m_2(x_2)$
			\\ 
			hierarchical interaction & 
			$m(x)=m_1(x_1)+m_2(x_2)+m_3(x_3)$
			\\ 
			+sparse &$ \qquad \qquad  +m_{1,2}(x_1,x_2)+m_{2,3}(x_2,x_3)$
			\\
			pure interaction+sparse & 
			$m(x)=m_{1,2}(x_1,x_2)+m_{2,3}(x_2,x_3)$
			\\
			additive+dense& 
			$m(x)=m_1(x_1)+\cdots+m_{d}(x_{d})$
			\\
			hierarchical interaction+dense& 
			$m(x)= \sum_{k=1}^{d}m_k(x_k) +   \sum_{k=1}^{d-1}m_{k,{k+1}}(x_k,x_{k+1})$
			\\
			pure interaction+dense& 
			$m(x)= \sum_{k=1}^{d-1}m_{k,{k+1}}(x_k,x_{k+1})$
			\\
			\hline
			\hline
			smooth& $m_k(x_k)=(-1)^{k}2\sin(\pi x_k)$\\
			& $m_{k,k+1}(x_k,x_{k+1})=m_k(x_kx_{k+1}) $
			\\
			jump & 
			$
			m_k(x_k)=
			\begin{cases}
			(-1)^{k}2\sin(\pi x_k) -2& \text{for}\ x\geq0,
			\\
			(-1)^{k}2\sin(\pi x_k) +2& \text{for}\ x<0
			\end{cases}
			$
			\\
			& $m_{k,k+1}(x_k,x_{k+1})=m_k(x_kx_{k+1}) $
			\\
			\hline \hline 
		\end{tabular}
    \end{table}
    The methods we compare are given in Table \ref{models2}. With sbf, gam and MARS we included the most popular methods used for models satisfying BI(1). We also included xgboost and random forest which are the most popular tree based methods used in practice. Furthermore, we included BART as a benchmark for methods not satisfying BI($r$) for small $r$. Note that BART has a high complexity and computation time due to the fact that trees are not only grown but also trimmed again in some iterations. 
	\begin{table}[ht]
		\centering
		\caption{Dictionary for algorithms. In total we consider 7 different ones. Additionally, 2 toy algorithms are considered for benchmark values.} \label{models2}
		\begin{tabular}{|lll|}
			\hline
			Description & short & Code-Reference\\ 
			\hline
			\hline
			A gradient boosting variant & 
			xgboost & \cite{Chen:2020aa}
			\\ 
			rpf & 
			rpf & 
   Supplement\\ 
			Random forest & rf & \cite{wright2015ranger}
			\\
			Smooth backfitting\footnote{Smooth backfitting for additive models with a local linear kernel smoother} & sbf & 
   Supplement
			\\
			Generalized additive models\footnote{A generalized additive model implementation via smoothing splines} & gam & \cite{wood2011fast}
			\\
			Bayesian additive regression trees & BART & \cite{sparapani2021nonparametric}
			\\
			multivariate additive regression splines & MARS & \cite{hastie2020package}
			\\
			\hline
			\hline
			1 nearest neighbours & 1-NN & \\
			Sample average & mean &
			\\
			\hline \hline 
		\end{tabular}
	\end{table}
	Performance is evaluated by the empirical mean squared error (MSE) from 100 simulations. The MSE is evaluated on test points $X_{501}^s,\dots,X_{1000}^s$ which are generated independently from the data
	\[
	    \frac{1}{100} \sum_{s=1}^{100}\frac{1}{500}\sum_{i=501}^{1000} \{m(X_i^s)-\widehat m^s(X_i^s)\}^2,
    \]
	where $\widehat{m}^s$ represents the estimator depending on the data $(X_i^s,Y_i^s)_{i=1}^{500}$. For xgboost and rpf we record results for the case when the estimators satisfy BI($1$). For the rpf algorithm, we further distinguish between an estimator satisfying BI($2$). We chose parameters via an independent simulation for each method from the parameter options outlined in Table \ref{parameters} beforehand. 
    For xgboost and rpf we also considered data-driven parameter choices (indicated by CV).
    We ran a 10-fold cross validation considering all parameters outlined in Table \ref{parameters} including all sub-methods and options.
    
	We only show selected parts of the overall study in the main part of this paper. Additional tables can be found in Section \ref{Section:Further Simulation Results} in the supplement.
	\begin{table}
        \centering
    	\caption{Model 1: Additive Sparse Smooth Model. We report the average MSE from 100 simulations. Standard deviations are provided in brackets.} \label{tab:1}
        \begin{tabular}{llllc}
            \hline
        	Method  & dim=4 & dim=10 & dim=30 & BI($r$) $r=0/1/2$ \\ 
            \hline
            xgboost (\texttt {depth=1})& 0.119 (0.021) & 0.142 (0.021) & 0.176 (0.027) & 1 \\ 
            xgboost  & 0.141 (0.024) & 0.166 (0.028) & 0.193 (0.033) &$\times$ \\ 
            xgboost-CV  & 0.139 (0.028) & 0.152 (0.029) & 0.194 (0.035) & $\times$\\ 
            rpf (\texttt {max\_interaction=1}) & 0.087 (0.018) & 0.086 (0.017) & 0.097 (0.019) & 1 \\ 
            rpf (\texttt {max\_interaction=2}) & 0.107 (0.015) & 0.121 (0.025) & 0.142 (0.026) & 2 \\ 
            rpf  & 0.112 (0.017) & 0.134 (0.026) & 0.162 (0.028) & $\times$\\ 
            rpf-CV& 0.103 (0.02) & 0.102 (0.035) & 0.105 (0.022) & $\times$\\ 
            rf & 0.209 (0.021) & 0.252 (0.027) & 0.3 (0.029) & $\times$\\ 
            sbf & 0.071 (0.026) & 0.134 (0.013) & 0.388 (0.073) & 1 \\ 
            gam  & 0.033 (0.012) & 0.035 (0.013) & 0.058 (0.021) &  1 \\ 
            BART  & 0.085 (0.019) & 0.076 (0.017) & 0.091 (0.023) & $\times$\\ 
            BART-CV& 0.09 (0.019) & 0.081 (0.014) & 0.09 (0.02) & $\times$\\ 
            MARS & 0.054 (0.014) & 0.061 (0.025) & 0.076 (0.031) & $\times$\\ 
            1-NN & 1.509 (0.1) & 3.228 (0.182) & 5.534 (0.313) &  $\times$\\ 
            average  & 3.811 (0.217) & 3.689 (0.183) & 3.748 (0.202) & 0 \\ 
            \hline
        \end{tabular}
    \end{table}
    Table \ref{tab:1} contains the additive sparse smooth setting. Considering algorithms satisfying BI($1$), we observe that algorithms relying on continuous estimators (sbf, gam) outperform the others with gam being the clear winner, irrespective of the number of predictors $d$. Rpf (\texttt {max\_interaction=1}) outperforms xgboost (\texttt {depth=1}). From the other algorithms, MARS does best with BART as runner up. Interestingly, rpf (\texttt {max\_interaction=2}) and rpf perform very well and even outperform xgboost (\texttt {depth=1}). In the hierarchical interaction sparse smooth setting which is visualized in Table \ref{tab:2}, we only show results from methods which can deal with interactions. Other cases are deferred to the supplement. Note that among the considered algorithms, rpf is the only algorithm satisfying BI(2).
    Performance wise, we find that BART as well as MARS outperform rpf. The latter slightly outperforms xgboost. In Table \ref{Pure Interaction-Sparse-Smooth} in the supplement, a similar picture can be observed in the pure interaction case. In particular, BART proves to be much stronger than all competing algorithms. This comes at the cost of increased computational time.
    
	Next, we re-visit the case where BI(1) is satisfied. However, this time the regression function is not continuous. The results are tabulated in Table \ref{tab:3}. Considering algorithms satisfying BI($1$), rpf (\texttt{max\_interaction=1}) performs best with xgboost (\texttt{depth=1}) following up. A visual picture of the excellent fit of rpf was provided in Figure \ref{fig:addjump}. From the other algorithms, BART is the first competitor.
	Unsurprisingly, models based on continuous estimators can not deal with jumps in the regression function. Cases including interaction terms and jumps are deferred to the supplement.  
	\begin{table}[ht]
		\centering
		\caption{Model 2: Hierarchical Interaction Sparse Smooth Model. We report the average MSE from 100 simulations. Standard deviations are provided in brackets.} 
		\label{tab:2}
		\begin{tabular}{llllc}
			\hline
			Method& dim=4 & dim=10 & dim=30 & BI($r$) $r=0/1/2$ \\ 
			\hline
            xgboost  & 0.374 (0.035) & 0.481 (0.064) & 0.557 (0.089) & $\times$ \\ 
            xgboost-CV & 0.393 (0.051) & 0.499 (0.058) & 0.563 (0.089) & $\times$ \\ 
            rpf (\texttt {max\_interaction=2}) & 0.248 (0.038) & 0.327 (0.045) & 0.408 (0.07) & 2 \\ 
            rpf  & 0.263 (0.034) & 0.357 (0.044) & 0.452 (0.076) & $\times$ \\ 
            rpf-CV  & 0.277 (0.039) & 0.366 (0.051) & 0.463 (0.083) & $\times$ \\ 
            rf &  0.432 (0.039) & 0.575 (0.061) & 0.671 (0.08) & $\times$ \\ 
            BART  & 0.214 (0.03) & 0.223 (0.04) & 0.252 (0.037) & $\times$ \\ 
            BART-CV  & 0.242 (0.043) & 0.276 (0.053) & 0.315 (0.047) & $\times$ \\ 
            MARS & 0.355 (0.089) & 0.282 (0.038) & 0.414 (0.126) & $\times$ \\ 
            1-NN& 2.068 (0.156) & 5.988 (0.624) & 11.059 (0.676) & $\times$ \\ 
            average & 8.366 (0.43) & 8.086 (0.246) & 8.207 (0.496) & 0 \\ 			
			\hline
		\end{tabular}
	\end{table}
	
	\begin{table}[ht]
		\centering
		\caption{Model 4: Additive Sparse  Jump Model. We report the average MSE from 100 simulations. Standard deviations are provided in brackets.} 
		\label{tab:3}
		\begin{tabular}{llllc}
			\hline
			Method & dim=4 & dim=10 & dim=30 & BI($r$) $r=0/1/2$ \\ 
	        \hline
            xgboost (\texttt {depth=1}) & 0.19 (0.029) & 0.282 (0.044) & 0.401 (0.045) & 1 \\ 
            xgboost & 0.198 (0.031) & 0.265 (0.053) & 0.286 (0.034) & $\times$\\ 
            xgboost-CV  & 0.209 (0.028) & 0.281 (0.052) & 0.313 (0.058) & $\times$\\ 
            rpf (\texttt {max\_interaction=1})& 0.159 (0.033) & 0.198 (0.075) & 0.179 (0.041) & 1 \\ 
            rpf (\texttt {max\_interaction=2}) & 0.185 (0.028) & 0.24 (0.066) & 0.259 (0.043) & 2 \\ 
            rpf &  0.192 (0.026) & 0.251 (0.065) & 0.282 (0.043) & $\times$\\ 
            rpf-CV &  0.169 (0.033) & 0.207 (0.072) & 0.183 (0.042) & $\times$\\ 
            rf &  0.274 (0.035) & 0.322 (0.05) & 0.375 (0.037) & $\times$\\ 
            sbf & 0.342 (0.049) & 0.603 (0.053) & 1.112 (0.138) & 1 \\ 
            gam  & 0.41 (0.047) & 0.406 (0.027) & 0.431 (0.06) & 1 \\ 
            BART & 0.177 (0.047) & 0.162 (0.038) & 0.157 (0.034) & $\times$\\ 
            BART-CV & 0.179 (0.051) & 0.163 (0.041) & 0.159 (0.036) & $\times$\\ 
            MARS  & 0.751 (0.136) & 0.74 (0.104) & 0.687 (0.123) & $\times$\\ 
            1-NN  & 2.393 (0.229) & 3.029 (0.308) & 3.512 (0.333) & $\times$\\ 
            average  & 1.276 (0.075) & 1.25 (0.063) & 1.213 (0.054) & 0 \\
            \hline
		\end{tabular}
	\end{table}
	
	We now consider dense models. 
	In the additive dense smooth model, see Table \ref{tab:4} in the supplement, the clear winners are sbf and gam, with gam having the best performance.
	This is not surprising, since these methods have more restrictive model assumptions than the others. Additionally, gam typically has an advantage over sbf in our setting. This is because splines usually fit trigonometric curves well, while kernel smoothers would be better off with a variable bandwidth in this case, which we did not implement. Observe that MARS does not deal well with this situation. Our rpf method is in third place, tied with BART and slightly outperforming xgboost. This suggests that the rpf is especially strong in sparse settings. In dense settings, the advantage towards xgboost and BART shrinks. This observation is underpinned in the  hierarchical interaction dense smooth model, tabulated in Table \ref{tab:5} in the supplement.
	While rpf and BART perform best with only four predictors, in dimension 10 xgboost outperforms rpf.

    We close this section by making some concluding remarks on the simulation results.
    Considering algorithms satisfying BI($1$), we observe that rpf (\texttt{max\_interaction}=1) adapts well to various models, in particular in sparse settings. While other algorithms such as gam outperform rpf in specific settings, rpf shows a good overall performance and higher flexibility. Additionally, rpf has the advantage of providing estimators which can approximate interactions and satisfy BI($r$) for predefined $r>1$. The simulations also show that non-restricted versions (rpf and rpf-CV) are competitive, while BART outperforms rpf in most cases. Note that BART is computationally quite intensive and we indeed struggled to get BART running in higher dimensional cases without error. Computational problems may increase if one goes to larger data sets, $n \gg 500$. Xgboost showed very strong results in terms of accuracy while being very fast and resources effective. However, accuracy turns out to be slightly worse than that of rpf in low-interaction settings.A short comparison of rpf-CV, xgboosting-CV, BART-CV and random forest can also be found in the supplement.

\section{Theoretical properties}\label{Theoretical properties in the additive case}

	In this section we derive asymptotic properties for a slightly modified rpf algorithm. The main difference is that in the tree  construction of the modified forest estimator, splitting does not depend on the responses $(Y_i: i=1,...,n)$ directly, at least not strongly, see Condition 2. With this modification we follow other studies of forest based algorithms to circumvent mathematical difficulties which
	arise if settings are analyzed where the same data is used to choose the split points as well as to calculate the fits in the leaves. Clearly, one can apply the results in this section to a similar modification of rpf making use of data splitting to separate splitting and fitting. An alternative route would be to use subsampling approaches as has been done in Mentch and Hooker (2016, 2017), Wager and Athey (2018), Peng et al. (2019). However, it is not clear if this would allow a detailed analysis of the rate of convergence of a subsampling version of rpf. Our results imply two findings. First for $r \leq 2$ the estimator can achieve optimal rates up to logarithmic terms in the nonparametric model where the interaction terms $m_t$ (for $|t|\leq 2$) allow for continuous second order derivatives. Secondly, for all choices of $r$ one achieves faster rates of convergence for the forest estimator than for tree  estimators that are based on calculating only one single tree. We will comment below why the situation changes for $r \geq 3$ compared to $r \leq 2 $. A major challenge in studying rpf lies in the fact that the estimator is only defined as the result of an iterative algorithm and not as the solution of an equation or of a minimizing problem. In particular, our setting differs from other studies of random forests where tree estimates are given by leaf averages of terminal nodes. In such settings  the tree estimator only depends on the terminal leaves, but not on other structural elements of the tree, and in particular not on the way the tree was grown. Secondly, the definition of the estimator as a leaf average allows for simplifications in the mathematical analysis.  The main point of our mathematical approach is to show that approximately the tree  estimators and the forest estimators are given by a least squares problem defined by the leaves at the end of the algorithm.
 
	We assume that the regression function fulfills BI($r$) with parameter $r\in\{1,\dots,d\}$. We write
	$m(x):=m^0(x) = \sum_{t \in T_r} m^0_t(x)$, where the components $m^0_t: t \in T_r$ are smooth functions which only depend on the values of $x_t$. The data is generated as 
	\begin{align*}
	    Y_i = m^0(X_i)+ \varepsilon_i = \sum_{t \in T_r} m^0_t(X_{i})+ \varepsilon_i.
	\end{align*}
	In the model equation, the $\varepsilon_i$'s are mean zero error variables and, for simplicity, the covariables $X_{i,k}$ are assumed to lie in $[0,1]$; $i=1,\dots,n; k=1,\dots,d$.
	We introduce a class of theoretical random planted forests. The class depends on an updating procedure of leaves $\mathcal{I}_{t,l}$ which must satisfy the conditions below. An example and explanation of the conditions is given in Section \ref{Example for a Theoretical Random Planted Forest Algorithm} in the supplement. As initialisation we set $\mathcal{I}^0_{t,1}= [0,1]^d$, $L_{t,0}=1$ and $\widehat m^{0}_{t } \equiv 0$ for $t \in T_r$. In iteration steps $s=1,...,S$, a leaf type $t_s\in T_r$ is selected and sets of a partition $\mathcal{I}^s_{t_s,l}=\prod_{k \in t_s} (a_{t_s,l,k}^s, b^s_{t_s,l,k}] \times \prod_{k \not \in t_s} (0,1]$ of $[0,1]^d$ with $l=1,...,L^s_{t_s}$ are updated with some (random) procedure satisfying the conditions below. Here and in the following, we sometimes write $f(x_t)$ instead of $f(x)$ for functions $f$ that depend only on $x$ via $x_t$. Similarly, we sometimes write $x_{t_s}\in I_{t_s,l}$. For a subset $\mathcal{I}$ of $[0,1]^d$ we write $|\mathcal{I}|$ for the Lebesgue measure of $\mathcal{I}$. Note that for finite sets $Q$, we also use $|Q|$ for the number of Elements in $Q$. Additionally, we write $|\mathcal{I}|_n=|\{i: X_i \in \mathcal{I}\}|/n$ for the empirical measure of $\mathcal{I}$. In order to update $\widehat m^{0}_{t }$, we define the density estimator $\widehat p^s_{t}(x_{t} ) = | \mathcal{I}^{s}_{t,l}|_n/| \mathcal{I}^{s}_{t,l}|$ for $x \in \mathcal{I}^{s}_{t,l}$ and the $|t\cup t'|$-dimensional estimator $\widehat p^s_{t\cup t'}(x_{t\cup t'} ) = | \mathcal{I}^{s}_{t,l}\cap  \mathcal{I}^{s}_{t',l'}|_n/| \mathcal{I}^{s}_{t,l}\cap  \mathcal{I}^{s}_{t',l'}|$ for $x \in \mathcal{I}^{s}_{t,l}\cap  \mathcal{I}^{s}_{t',l'}$. With this notation an update for leave type $t_s \in T_r$ can be written as
	\begin{eqnarray} 
	    &&\widehat m_{t_s} ^s (x_{t_s} ) = \widehat{  m}_{t_s} ^{*,s} (x_{t_s} ) - \sum _{t \in T_r, t \not = t_s}\int_{(0,1]^{|t\backslash t_s|}}
	    \frac {\widehat p^s_{t_s,t}(x_{t_s}, u_{t \backslash t_s} )}   {\widehat p^s_{t_s}(x_{t_s} )} 
	    \widehat m^{s-1}_{t } (x_{t \cap t_s}, u_{t \backslash t_s} ) \mathrm d u_{t \backslash t_s}. \label{eq:backf}	
	\end{eqnarray}
	Here for $x_{t_s} \in \mathcal{I}^{s}_{t_s,l} $ with $l \in \{1,...,L_{t_s}^s\}$ the function
	$\widehat{ m}_{t_s} ^{*,s}$ is a marginal estimator defined by $\widehat{  m}_{t} ^{*,s} (x_{t} ) = \frac 1 {n|\mathcal{I}^{s}_{t,l}|_n} \sum _{X_i \in \mathcal{I}^{s}_{t,l}} Y_i$. 
    For $t \not = t_s$ we set $\widehat m^{s}_{t } =\widehat m^{s-1}_{t }$. Note that \eqref{eq:backf} is a mathematically convenient rewriting of an update in the rpf algorithm. For $l \in \{1,...,L_{t_s}^s\}$ residuals for $Y_i$ with $X_{i,t_s} \in \mathcal{I}^{s}_{t_s,l} $ are averaged and the average is added to the old fit 
    of $m$ in 
    $\mathcal{I}^{s}_{t_s,l} $.
    After $S$ ($=\texttt{nsplits}$) steps we get the estimators $\widehat m_{t} ^S$ for $t \in T_r$. 
    Now, the tree  estimator of the function $m^0$ is given by $\widehat m^S(x) = \sum_{t \in T_r} \widehat m^S_t(x_t)$. For the forest estimator tree  estimators are repeatedly constructed. They are denoted by $ \widehat m^{S,v}_t$ for $t \in T_r$ and $ \widehat m^{S,v}$ for $v=1,...,V$ ($=\texttt{ntrees}$). We define the forest estimator as $\widehat m_t (x_t)= V^{-1} \sum_{v=1} ^V\widehat m^{S,v}_t (x_t)$ for $t \in T_r$ and $\widehat m (x) = V^{-1} \sum_{v=1} ^V\widehat m^{S,v}(x)$. 
    If necessary, we will also write $\mathcal{I}^{s,v}_{t,l}$, $a^{s,v}_{t,l,k}$, $t^v_s$ ,$k^v_s$, ... instead of $\mathcal{I}^{s}_{t,l}$, $a^{s}_{t,l,k}$, $t_s$ ,$k_s$, ... to indicate that we discuss the tree  estimator with index $v$.

        \subsection{Main results}

For our results we make use of the following assumptions.
    
	\begin{condition} The tuples $(X_i, \varepsilon_i)$ are i.i.d.
	The functions $m^0_t$ are twice continuously differentiable and $E[m^0_t(X_{i,t})]=0$ for all $t\in T_r$. The covariable $X_i$ has a density $p$ that is bounded from above and below (away from $0$). For $t\in T_{2r}$ the  joint density $p_{t}$ of the tuple $X_{i,t}$ allows  continuous derivatives of order 2.  Conditionally on $X_i$ and the iterative construction of the leaves, the error variables $\varepsilon_i$ have mean zero, variance bounded by a constant, and the products $\varepsilon_i \varepsilon_j$ are mean zero for $i \not = j$. Conditionally on $X_i$ the iterative construction of the leaves in the trees are i.i.d. for $v= 1, ..., V$.
	\end{condition}
 
    Note that the number $S$ of iterations in the tree construction and the number $V$ of constructed trees may depend on $n$. The parameter $S$ is of the same order as the number of leaves in the final partition. $S$ is assumed to converge to infinity, see Condition 3. When considering the forest estimator one usually requires $V$ to converge to infinity as well in order to obtain useful convergence rates, see also Conditions 7,8. In the next condition we assume that each leave type is chosen often enough.
	
	\begin{condition} For  $C_{A3} > 0$ large enough we assume that there exists a constant $C^\prime_{A3} > 0$ such that, with probability tending to one, for $\{ s : S-J' \leq s \leq S\}$ and $1 \leq v \leq V$ there exists a partition $S-J' = s_0^v < s_1^v < ...< s_J^v=S$ such that for $1 \leq j \leq J$ the set $\{t_s^v: s_{j-1}^v< s \leq s_j^v\}$ contains all elements of $T_r$. Here $J$ and $J'$ are the smallest integers larger or equal to $C_{A3} \log n$ and $C^\prime_{A3} (\log n)^2$ respectively.
	\end{condition}
 
We now assume that the tree partitions become fine enough. 

   \begin{condition}	For $x \in [0,1]^d$, $t\in T_r$, $1 \leq v \leq V$ we define $l_t(x)=l^v_t(x)$ such that $x \in \mathcal{I}^{S,v}_{t,l^v_t(x)}$. We assume that  with probability tending to one uniformly over $t\in T_r$, $1 \leq v \leq V$, $k\in t$
    \begin{eqnarray*}  
        &&\frac 1 n \sum_{i=1} ^n \left(b^{S,v}_{t,l^v_t(X_{i}),k}- a^{S,v}_{t,l^v_t(X_{i}),k}\right)^2 \leq \delta^2_{1,n}
    \end{eqnarray*}  
    for a sequence $ \delta_{1,n}$ with  $(\log n)^2 \delta_{1,n} \to 0$ and $n^R \delta_{1,n} \to \infty $ for $R>0$ large enough.
	\end{condition}

The next two conditions assume that there are not too large changes in the tree partitions in the last logarithmic iterations and that the histogram estimators in the updating equations converge to the underlying design densities.

    \begin{condition}	It holds for $t,t' \in T_r $, $1 \leq v \leq V$ and $S- J' \leq s \leq S$ that
	\begin{eqnarray*}
	    \sup _{t, t' \in T_r} \int _{(0,1] ^{|t\cup t'|}} \left ( \frac {\widehat p^{S,v}_{t\cup t'}(u_{t\cup t'})} {\widehat  p^{S,v}_{t}(u_t)} -\frac {\widehat p^{s,v}_{t\cup t'}(u_{t\cup t'})} {\widehat p^{s,v}_{t}(u_t)}
	    \right )^2 \mathrm d u_{t\cup t'} &\leq& \delta^2_{1,n} (\log n) ^{-4} , \\
	    \sup _{t \in T_r} \int _{(0,1] ^{|t|}} \left (   \widehat{  m}_{t} ^{*,S,v} (u ) -   \widehat{  m}_{t} ^{*,s,v} (u )
	    \right )^2 \mathrm d u &\leq& \delta^2_{1,n}  (\log n) ^{-4}
    \end {eqnarray*}
	with probability tending to one.
	\end{condition}

	
	\begin{condition}	
    	It holds uniformly  for $t,t' \in T_r $, $1 \leq v \leq V$, $s=S,S-J'-1$  that 
    	\begin{eqnarray*}
    	    \int _{(0,1] ^{|t\cup t'|}} \left ( \frac {\widehat p^{S,v}_{t\cup t'}(u_{t\cup t'})} {\widehat  p^{S,v}_{t}(u_t)} -\frac { p_{t\cup t'}(u_{t\cup t'})} { p_{t}(u_t)}
    	    \right )^2 \mathrm d u_{t\cup t'} &\leq& \delta^2_{2,n},\\ \sup _{u \in (0,1]^{|t|}} \left | \widehat p^{s,v}_{t}(u) - p_{t}(u) \right |   &\leq& \eta_{1,n}
    	\end{eqnarray*} 
    	 with probability tending to one for sequences $\delta_{2,n}$, $\eta_{1,n}$ with  $(\log n)^2 \delta_{2,n} \to 0$ and $ \eta_{1,n} \to 0$.
	\end{condition}

	\begin{theorem}\label{theomain1}
	    Under Conditions $2,\dots,6$, for $1 \leq v \leq V$ the tree  estimators satisfy
	    \begin{eqnarray*}  
	        \left \| \sum_{t \in T_r}(\widehat m^{S,v}_t- m^0_t) \right \| &=& O_P(\delta_{1,n} + S^{1/2} n^{-1/2})
		\end{eqnarray*}
		with probability tending to one, where $\|\cdot\|$ denotes the $L_2(P)$ norm.
	\end{theorem}
	
	The proof of Theorem \ref{theomain1} can be found in Section \ref{prooftheo1}. We shortly discuss this result. Suppose that the leaves $\mathcal{I}^{s}_{t,l}  $ have side lengths of order $h$ for some sequence $h \to 0$. Then $\delta_{1,n}$ is of order $h$, $S$ is of order $h^{-r}$ and up to logarithmic terms we obtain a rate of order $h + (nh^r)^{-1/2}$ for the estimation error of $\widehat m^{S,v} = \sum_{t \in T_r}\widehat m^{S,v}_t$.

 We now discuss the performance of the forest estimator. For the result we need the following additional assumptions for discussing the complete forest estimator.
    \begin{condition}	
        It holds uniformly  for $t,t' \in T_r $, $1 \leq v \leq V$  that 
    	\begin{eqnarray*}
    	    \sup _{u \in (0,1]^{|t\cup t'|}} \left | \widehat p^{S,v}_{t\cup t'}(u) - p_{t\cup t'}(u) \right |   &\leq& \eta_{2,n},\\
    		\| \widehat p^{S,v}_{t} - p_{t} \|, \left \| \int_{(0,1]^{|t' \backslash t| } } \frac {
    	    \widehat p^{S,v} _{t\cup t'}(\cdot,u_{t'\backslash t}) -p_{t\cup t'} (\cdot, u_{t'\backslash t})}	{ p_{t}(\cdot)}   m^0_{t'} (\cdot, u_{t'\backslash t}) \mathrm d u_{t' \backslash t} 
    	    \right \| &\leq& \delta_{3,n}
    	\end{eqnarray*} 
    	with probability tending to one for some sequences $ \eta_{2,n}, \delta_{3,n} \to 0$.
	\end{condition}

	\begin{condition}
    	For $t,t' \in T_r $ we write   $\widehat p^{S,+}_{t\cup t'}= V^{-1} \sum_{v=1} ^V\widehat p^{S,v}_{t\cup t'}$ and $\widehat p^{S,+}_{t}= V^{-1} \sum_{v=1} ^V\widehat p^{S,v}_{t}$. It holds uniformly for $t,t' \in T_r $ that 
    	\begin{eqnarray*}
    		\| \widehat p^{S,+}_{t} - p_{t} \|_{1}, \left \| \int_{(0,1]^{|t' \backslash t| } } \frac {
    	    \widehat p^{S,+} _{t\cup t'}(\cdot, u_{t'\backslash t}) -p_{t\cup t'} (\cdot, u_{t'\backslash t})}	{ p_{t}(\cdot)}   m^0_{t'} (\cdot, u_{t'\backslash t}) \mathrm d u_{t' \backslash t} 
    	    \right \|_{1} &\leq& \delta_{4,n}
    	\end{eqnarray*} 
    	with probability tending to one for a sequence $\delta_{4,n} \to 0$. Here $\|\cdot\|_{1}$ denotes the  $L_1(P)$ norm.
	\end{condition}

    Below we will argue that the averaged estimators $\widehat p^{S,+}_{t\cup t'}$ and $\widehat p^{S,+}_{t} $ in Condition 8 can achieve rates of convergence $\delta_{4,n}=O(\delta^2_{1,n})$ compared to their summands which have a bias of order $\delta_{1,n}$. If the density $p_t$ is twice continuously differentiable,  qualitatively, $\widehat p^{S,+}_{t}$ behaves like a kernel density estimator with bandwidth $h$ of order $\delta_{1,n}$. We switch from the $L_2(P)$ norm to the $L_1(P)$ norm in  Condition 8 and the next theorem. The reason is that at the boundary of size $Ch$ with $C$ large enough we have a bias of order $h$ where the bias is of order $h^2$ in the interior. Thus measured by the $L_1(P)$ norm we get a bias of order $h^2$ whereas the $L_2(P)$ norm has a slower rate caused by the boundary effects.\\ 

	\begin{theorem}\label{theomain2} 
	    Under Conditions $2,\dots,8$, for the forest estimator we have
	    \begin{eqnarray*}
	        &&\left \| \sum_{t \in T_r}(\widehat m_t- m^0_t) \right \|_{1} \leq C 
         ( \delta_{1,n}^2  + \delta_{1,n}\delta_{2,n}
 + \delta_{3,n}^2 + \delta_{4,n}+ S^{1/2} n^{-1/2})
	    \end{eqnarray*} 
	    with probability tending to one.
	\end{theorem}
	
	The proof of the theorem will be given in Section \ref{prooftheo2}. Again, we shortly discuss this result. As above, suppose that the leaves $\mathcal{I}^{s}_{t,l}$ have side lengths of order $h$ for some sequence $h \to 0$. Then $\delta_{1,n}$ is of order $h$ and $S$ is of order $h^{-r}$. The sequence $\delta_{2,n}$ is the rate of a histogram estimator of dimension $\leq 2r$ which up to logarithmic terms is of order $h + (nh^{2r})^{-1/2}$ for $2r$ dimensional estimators. For $x\in[0,1]^{|t|}$ sufficiently far away from the boundary of $[0,1]^{|t|}$ assume that the random variables $x_k-a_k^S$ and $b_k^S-x_k$ approximately follow the same distribution, where $a_k^S,b^S_k$ are the lower and upper bounds of the leaf which contains $x$ with respect to dimension $k$. Then   the bias terms of order $h$ cancel and we get that the bias terms measured by $\delta_{4,n}$ are of order $h^2$.  Thus, up to logarithmic terms, we get a bound on the accuracy of the forest estimator of order $h^2 + h  (nh^{2r})^{-1/2} +  (nh^{r})^{-1/2}$. This rate is faster than the tree  rate if $nh^{2r}\to \infty$, i.e. we need consistency of $2r$ dimensional histogram estimators. Let us discuss this for a bandwidth $h$ that is rate optimal for the estimation of twice differentiable functions with $r$ dimensional argument. Then $h$ is of order $n^{-1/(r+4)}$ and the $2r$-dimensional estimators are consistent for $r \leq 3$. Thus, we get from our theory that for optimal tuning parameters forest estimators outperform tree estimators. For $r\leq 2$ we obtain optimal rates by the forest estimator, i.e. $n^{-2/5}$ for $r=1$ and $n^{-1/3}$ for $r=2$.
	
	We now explain where the consistency of the density histogram estimators is essential. The  estimators show up as kernels in integral equations which define the tree and forest estimators. This means that, approximately, the tree  estimators $\widehat m^{v}$ are  given as solutions  of integral equations of the form $\widehat m^{v} = \bar m^{v} + A^v \widehat m^{v}$ with random integral operators $A^v$, where the operators $A^v$ are defined by up to $2r$ dimensional density estimators. If these density estimators are consistent, the operators $A^v$ are approximately equal to an operator $A$ not depending of $v$. Then $\widehat m^{v}$ approximately solves $\widehat m^{v} = \bar m^{v} + A \widehat m^{v}$ and
    the forest estimator $\widehat m =V^{-1} \sum_{v=1} ^V\widehat m^{v}$ approximately solves $\widehat m = V^{-1} \sum_{v=1} ^V \bar m^{v} + A \widehat m$. In order to get faster convergence rates for the forest estimator one shows that the average $V^{-1} \sum_{v=1} ^V \bar m^{v}$ has faster rates as the summands $\bar m^{v}$. 
    
    For the whole argument it is crucial that up to $2r$ dimensional density estimators are consistent. In case these estimators are inconsistent invertibility of the operator $I - A$ does not carry over to the operator $I - A^v$. Thus, the equation $\hat{m}^v=\hat{m}^v + A^v\hat{m}^v$ may not have a unique solution $\hat{m}^v$ which suggests that the solution to which the algorithm converges depends on how the tree was grown. In particular, this excludes a mathematical study along our lines.

 \subsection{Proof of Theorem \ref{theomain1}} \label{prooftheo1} In the proof we denote different constants by  $C$. The meaning of $C$ may change, also in the same formula.  In this proof we omit the index $v$ in the notation.
	
	 We rewrite \eqref{eq:backf} as 
	\begin{eqnarray} \widehat m ^s (x ) &=& \widehat{  m} ^{*,s} (x_{t_s} )  + \widehat \Psi^s \widehat m ^{s-1}  (x ), \label{eq:backf2}	\end{eqnarray}  
	where $ \widehat{  m}^{*,s}= \widehat{  m}_{t_s} ^{*,s}$, $ \widehat \Psi^s  =  \widehat \Psi_{t_s}^s $ and where 
	\begin{eqnarray*} &&\widehat \Psi_t^s m (x) = m(x) - m_{t} (x_{t}) \\	 
	  &&\qquad
	- \sum _{t' \in T_r, t' \not = t}\int_{(0,1]^{|t'\backslash t|}}
	 \frac {\widehat p^s_{t\cup t'}(x_{t}, u_{t '\backslash t} )}   {\widehat p^s_{t}(x_{t} )} 
	 m_{t '} (x_{t' \cap t}, u_{t '\backslash t} ) \mathrm d u_{t' \backslash t} \end{eqnarray*}
	for a function $m(x) = \sum_{t \in T_r} m_t(x_t)$. Iterative application of \eqref{eq:backf2} gives for $1 \leq S_1 < S_2 \leq S$
	 \begin{eqnarray} \widehat m ^{S_2}  &=& \widehat{  m} ^{*,S_2}  + \widehat \Psi^{S_2}  \widehat{  m} ^{*,S_2-1}+ \widehat \Psi^{S_2}\widehat \Psi^{S_2-1}  \widehat{  m} ^{*,S_2-2}  \nonumber \\
	 && \qquad + ...+  \widehat \Psi^{S_2}...\widehat \Psi^{S_1+1}  \widehat{  m} ^{*,S_1}  +   \widehat \Psi^{S_2}...\widehat \Psi^{S_1}  \widehat{  m} ^{S_1-1} . \label{eq:backf3}	\end{eqnarray}  
	  We will use this equality in the proofs of the following lemmas for the choices $S_1 = 1 , S_2 = S^*$ and $S_1 = S^* , S_2 = S$ with $S^* = S - J'$.
	 
	\begin{lemma}\label{lem1} It holds  that
	\begin{eqnarray*}\frac 1 n \sum_{i=1} ^n  \widehat m ^{S^* -1} (X_i)^2&\leq & S^{1/2}  \frac 1 n \sum_{i=1} ^n  Y_i^2
			.\end{eqnarray*}
	\end{lemma}
	
	\begin{proof}
	    It can be easily shown that for $t \in T_r$
    	\begin{eqnarray}
    	    \frac 1 n \sum_{i=1} ^n  \widehat {\bar m} _t (X_i)^2 \leq \frac 1 n \sum_{i=1} ^n  Y_i^2.\label{eq:backf4}	
    	\end{eqnarray} 
    	Furthermore, for any function $m$ we have
    	\begin{eqnarray}
    	    \frac 1 n \sum_{i=1} ^n  \left (\widehat \Psi^s  m\right ) (X_i)^2 \leq \frac 1 n \sum_{i=1} ^n  m (X_i)^2.\label{eq:backf5}
    	\end{eqnarray} 
    	For a proof of \eqref{eq:backf5} consider the space of functions $m:[0,1]^d \to \mathbb{R}$ endowed with the pseudo metric 
    	$$ \|m\|_{n,2}^2 =  \frac 1 n \sum_{i=1} ^n  { m} (X_i)^2.$$ Then $I - \widehat \Psi^s$ is the orthogonal projection onto the space of piecewise constant functions that are constant on the leaves  $\mathcal{I}^s_{t,l}: l=1 ,...,L_{t}^s$. Furthermore, $ \widehat \Psi^s$ is the orthogonal projection onto the orthogonal complement of this  space. This shows \eqref{eq:backf5}. The statement of the lemma now follows from \eqref{eq:backf4} and \eqref{eq:backf5} by application of  \eqref{eq:backf3} with $S_1 =1$ and $S_2 = S^* -1$.
	\end{proof}
	
	The previous lemma can be used to show a bound for the $L_2(p)$ norm of $ \widehat m ^{S^* -1}$ which we denote by  $\| \widehat m ^{S^* -1}\|$
	
	\begin{lemma}\label{lem2} 
	    It holds  that with probability tending to one
	    \begin{eqnarray*}
	        \| \widehat m ^{S^* -1}\| ^2 =\int \widehat m ^{S^* -1} (x)^2 p(x) \mathrm d x &\leq & (1 -  C\eta_{1,n})^{-1}  S^{1/2}  \frac 1 n \sum_{i=1} ^n  Y_i^2
			.
		\end{eqnarray*}
	\end{lemma}
	
	\begin{proof}
	    For $t \in T_r$ we get by application of Condition 6 that
    	\begin{eqnarray*} 
    	    \int \widehat m ^{S^* -1}_t (x_t)^2 p(x) \mathrm d x &\leq& \int \widehat m ^{S^* -1}_t (x_t)^2 \widehat p^{S^* -1}_t(x_t) \mathrm d x_t\\
    	    && \qquad  + \int \widehat m ^{S^* -1}_t (x_t)^2 |\widehat p^{S^* -1}_t -   p_t|(x_t) \mathrm d x_t\\
    	    &\leq&\frac 1 n \sum_{i=1} ^n  \widehat { m} ^{S^* -1}_t (X_{t,i})^2\\
    	    && \qquad  +   C\eta_{1,n} \int \widehat m ^{S^* -1}_t (x_t)^2p_t(x_t) \mathrm d x_t.	
        \end{eqnarray*}
    	The statement of the lemma now follows by application of Lemma \ref{lem1}.
	\end{proof} 
	
	\begin{lemma}\label{lem3} 
	    It holds for some $C>0$ and $0 < \rho < 1$  that with probability tending to one
	    \begin{eqnarray*}\| 
	        \widehat \Psi^{S}\circ ... \circ\widehat \Psi^{S^*} \| \leq C \rho^{C_{A3} \log n}.
	    \end{eqnarray*}
	\end{lemma}

    Here for an operator $\Psi$ mapping $L_2(p)$ into $L_2(p)$ we denote the operator norm $ \|\Psi\| = \sup\{\|\Psi m\|: m \in  L_2(p), \|m\| =1\}$ by $ \|\Psi\| $.
    Before we come to the proof of this lemma let us discuss its implications. Using the representation \eqref{eq:backf2} we get from Lemmas  \ref{lem2},  \ref{lem3} that for all $R>0$ we can choose a constant $C_R$ such that if $C_{A3}$ is large enough with probability tending to one
    \begin{eqnarray}
        \| \widehat m ^{S} - \widehat m ^{1,S}\| \leq C_R n^{-R}, \label{eq:backf6}
    \end{eqnarray} 
	where
    \begin{eqnarray*} 
        \widehat m ^{1,S} &=& \widehat{  m} ^{*,S}  + \widehat \Psi^{S}  \widehat{  m} ^{*,S-1}+ ...+  \widehat \Psi^{S}...\widehat \Psi^{S^*+1}  \widehat{  m} ^{*,S^*}  . 
    \end{eqnarray*}  	
    Note that $\widehat m ^{1,S}$ only depends on the growth history of the tree in the last $C^{\prime}_J (\log n)^2$ steps. Thus we have shown that approximately the same holds for the tree estimators  $\widehat m ^{S}$. Below we will go a step further and show that the tree  estimator approximately only depends on the data averages in the terminal leaves and in particular not on the growth of the tree  in the past. Before we come to this refinement we first give a proof of Lemma \ref{lem3}. For this purpose we will introduce population analogues $ \Psi_{t_s}$ of the operators $ \widehat \Psi^{s}$ and  we will discuss some theory on backfitting estimators in interaction  models. We consider the following subspaces of functions $m: [0,1]^d \to  \mathbb{R}$ with $ E[m^2(X_i)] < \infty$ for $t \in T_r$
	\begin{eqnarray*}
		\mathcal H&=& \{ m: [0,1]^d \to  \mathbb{R}\ |\ E[m^2(X_i)] < \infty\}, \\
		\mathcal H_t&=& \{ m \in \mathcal H\ |\ m(x) = m_t(x_t) \text{ for some function } m_t: [0,1]^{|t|} \to  \mathbb{R}\},\\
		\mathcal H_{add}&=& \{ m \in \mathcal H\ |\ m(x) = \sum_{t \in T_r}m_t(x_t) \text{ for some functions }  m_t \in \mathcal H_t, t \in T_r\}.
	\end{eqnarray*}
	In particular, $\mathcal H_{\emptyset}$ is the subspace of $\mathcal H$ that contains only constant functions. 
	In abuse of notation  for a function $m_t \in \mathcal H_t$ we also write $m_t(x)$ with $x \in [0,1]^d$ instead of $m_t(x_t)$. Thus $m_t$ can be interpreted as a function with domain $[0,1]^{|t|} $ or with domain $[0,1]^d$.
	The projection of $\mathcal H$ onto $\mathcal H_{t}$ is denoted by $\Pi_t$. For $\Psi_t = I - \Pi_t$ and $m=  \sum_{t \in T_r}m_t$ with $m_t \in \mathcal H_{t}$ $( t \in T_r)$ one gets
	$$\Psi_t m (x) = \sum_{w\in T_r \backslash \{t\}} m_w(x_w) + m_t^*(x_t)  $$
	with
	$$m_t^*(x_t) = - \sum_{w \in T_r, w \not = t} \int { \frac {p_{t\cup w}(x_{t\cup w})}{p_{t}(x_t)}m _{w}(x_{w}) \mathrm d x_{w\backslash t}}.$$
	We now consider operators $K$ of the form $\Psi_{t^1}\circ \ldots \circ \Psi_{t^{k_K}}$ with $\{t^1,...,t^{k_K}\} \supseteq  T_r$. We call these operators complete. We argue that  there exists a constant $\gamma < 1$ such that for all complete operators $K$ of this form we have
    \begin{equation}
        \label{eq:Kbound} \| K \|_{add} = \sup \{ \| K m \| : m \in \mathcal H_{add}, \|m\| \leq 1 \} < \gamma.
    \end{equation}
	Note that in particular $\gamma$ does not depend  on the order of $t^1,...,t^{k_K}$. Our notation is a little bit sloppy because in the representation $\sum_{t \in T_r} m_t$ of $m$ the summands are not uniquely defined because $t\cap t'$ may be nonempty for some $t,t' \in T_r$, $t \not = t'$. A more appropriate notation would be to define $K$ as an operator mapping $\prod_{t\in T_r} \mathcal H_{t}$ into $\prod_{t\in T_r} \mathcal H_{t}$ and to endow this space with the pseudo norm $\|\sum_{t \in T_r}  m_{t}\|$. For a proof of \eqref{eq:Kbound} we will show that  $\sum _{t \in T} \mathcal H_{t}$  are closed subspaces of $\mathcal H$ for all choices of $T \subset T_r$. In particular, by a result of Deutsch (1985), see also Appendix A.4 in Bickel, Klaassen, Ritov and Wellner (1993), this implies that $\rho( \mathcal H_{t_j},  \mathcal H_{t_j+1}+ ...+  \mathcal H_{t_{k^K}} )<1$ for $1 \leq j \leq t_{k^K-1}$, where for two linear subspaces $L_1$ and $L_2$ of $\mathcal H$ the quantity $\rho(L_1.L_2)$ is the cosine of the minimal angle between $L_1$ and $L_2$, i.e. 
	\[
	    \rho(L_1.L_2) = \sup \left\{ \int h_1(x) h_2(x) p(x) \mathrm d x: h_j \in L_j \cap (L_1 \cap L_2)^\bot, \int h_j^2(x) p(x) \mathrm d x \leq 1,\ j=1,2\right\}.
	\]
	According to a result of Smith, Solomon and Wagner (1977) this implies that for an operator $K$ of the above form we have that $$ \| K \|_{add} \leq 1 - \prod_{j=1} ^{k^K} \sin^2(\alpha_j),$$ where $\alpha_j$ is chosen such that $\cos(\alpha_j)  = \rho( \mathcal H_{t_j},  \mathcal H_{t_{j+1}}+ ...+  \mathcal H_{t_{k^K}} )$. We remark that this bound is also valid if the space $\mathcal H_{t_j}$ is identical to the same space $\mathcal H_{t}$ for several choices of $j$. In this case we have $\sin^2(\alpha_j)=1$ for all such values of the index $j$ with the exception of the last appearance of $\mathcal H_{t}$. Because there are only finitely many ways to order $|T_r|$ elements we get by the last remark that \eqref{eq:Kbound} holds with some $\gamma < 1$ for all operators $K$. For a proof of \eqref{eq:Kbound} it remains to show that  $\sum _{t \in T} \mathcal H_{t}$  are closed subspaces of $\mathcal H$ for all choices of $T \subset T_r$. For this claim we will argue that $\sum _{t \in T} \bar {\mathcal H_{t}}$  are closed subspaces of $\mathcal H$ for all choices of $T \subset T_r$, where $ \bar {\mathcal H_{t}} = \{ h \in  {\mathcal H_{t}}: \int h(x) \mathrm d x_j=0 $ for $ j\in t\}$. According to Proposition 2 in the supplement material A.4 of Bickel, Klaassen, Ritov and Wellner (1993) this follows if there exists some $c> 0$ such that $\int (\sum_{t \in T} h_t(x_t) )^2 p(x) \mathrm d x \geq 1 $ implies that  $\int h_t(x_t) ^2 p(x) \mathrm d x \geq c $ for some $t \in T$. This can be easily verified with $c = C(\max_{x \in [0,1]^d} p(x) ) ^{-1} |T_r|^{-1}$ by noting that for $h_t \in \bar {\mathcal H_{t}}$ it holds that
	\begin{eqnarray*}
	    \int \left(\sum_{t \in T} h_t(x_t) \right)^2 p(x) \mathrm d x &\leq& \max_{x \in [0,1]^d} p(x) \int \left(\sum_{t \in T} h_t(x_t) \right)^2 \mathrm d x\\ 
	    &=&\max_{x \in [0,1]^d} p(x)   \sum_{t \in T}\int h_t(x_t) ^2 \mathrm d x\\
	    &\leq& C \max_{x \in [0,1]^d} p(x) \sum_{t \in T}\int h_t(x_t) ^2p(x) \mathrm d x
    \end{eqnarray*}
	In particular, one can use  \eqref{eq:Kbound} to show that with $$\bar m_t(x) = E[ Y_i| X_{i,t}=x_t] = \sum_{t' \in T_r}  \int  m_{t'}(x_{t'}) \frac {p_{t'\cup t} (x_{t'\cup t} ) } {p_{ t} (x_{t} ) } \mathrm d x_{t' \backslash t}$$ we have  for $\mu^* \in \mathcal H_{add}$ with some $0 < \gamma < 1$, and some $C> 0$
	\begin{eqnarray*}
	    \| m -  \mu\|&\leq & C \gamma^{S^{*}} (\| \mu^*\|+1),
	\end{eqnarray*}
	where 
	\begin{eqnarray*}
		\mu = \bar m_{t^*_{v,1}} + \Psi_{t^*_{v,1}} \bar m_{t^*_{v,2}}
		+ \Psi_{t^*_{v,1}} \Psi_{t^*_{v,2}}  \bar m_{t^*_{v,3}}+ ...+  \Psi_{t^*_{v,1}} \circ ... \circ \Psi_{t^*_{v,S^{*}-1}} \mu^*
	\end{eqnarray*}
	with $t^*_{v,s} = t_{v,S-s+1}$. Note that $$m -  \mu = \Psi_{t^*_{v,1}} \circ ... \circ \Psi_{t^*_{v,S^{*}-1}} (m - \mu^*).$$
		
	\begin{proof}[Proof of Lemma \ref{lem3}]
	    By Condition 6 we have by application of Cauchy-Schwarz inequality that 
		\begin{eqnarray} \label{eq:boundpsipsihat} \|  \widehat \Psi^{s} - \Psi_{t_s} \| \leq C \delta_{2,n} \end{eqnarray}
		with probability tending to one. This implies that, with probability tending to one,
		$$\|  \widehat \Psi^{s} \| \leq 1 + C \delta_{2,n}$$
		and we get by a telescope argument  that, with probability tending to one, for $1 \leq j \leq J \leq C_{A3} \log n +1$
		\begin{eqnarray*} 
		    && \|\widehat \Psi^{s_{j}}\circ ... \circ\widehat \Psi^{s_{j-1} +1} -  \Psi_{t_{s_{j}}}\circ ... \circ \Psi_{t_{s_{j-1}  +1}} \| \leq C J^\prime (1 + C \delta_{2,n})^{J^\prime-1} \delta_{2,n} \\
		    && \qquad \leq C C_{A3}^\prime (\log n)^2 (1 + C \delta_{2,n})^{C^\prime_{A3} (\log n)^2} \delta_{2,n} \leq C  (\log n)^2  \delta_{2,n}.
		\end {eqnarray*}
		Because by Condition 3 $\Psi_{t_{s_{j}}}\circ ... \circ \Psi_{t_{s_{j-1}+1}}$ is a complete operator with high probability, we get from \eqref{eq:Kbound} that,  with probability tending to one, for $1 \leq j \leq J \leq C_{A3} \log n +1$ 
        \begin{eqnarray*}
		    \|\widehat \Psi^{s_{j}}\circ ... \circ\widehat \Psi^{s_{j-1}  +1} \|		&\leq& \gamma + C ( \log n)^2  \delta_{2,n}.
		\end {eqnarray*}
		This inequality can be used to show the bound on $\| \widehat \Psi^{S}\circ ... \circ\widehat \Psi^{S^*} \| $, claimed in Lemma \ref{lem3}.
	\end{proof}  

    As discussed above, Lemma \ref{lem3} implies \eqref{eq:backf6}. We now approximate $\widehat m ^{1,S}$ by $\widehat m ^{2,S}$ where
    \begin{eqnarray*} 
        \widehat m ^{2,S} &=& \widehat{  m} ^{*,S} _{t_{S}} + \widehat \Psi_{t_S}^{S}  \widehat{  m} ^{*,S} _{t_{S-1}}
        + \widehat \Psi_{t_S}^{S} 
        \widehat \Psi_{t_{S-1}}^{S}
        \widehat{  m} ^{*,S}_{t_{S-2}}
        + ...+  \widehat \Psi_{t_S}^{S}...\widehat \Psi_{t_{S^*+1}}^{S}  \widehat{  m} ^{*,S}_{t_{S^*}}  . 
    \end{eqnarray*}  
    Note that  $\widehat m ^{2,S}$ differs from  $\widehat m ^{1,S}$ by having always the superindex $S$ for the operators $\Psi$ and the functions $\widehat{  m}^* $. In the following lemma we compare  $\widehat m ^{S} $ and $\widehat m ^{2,S}$. The  bound can be shown by similar arguments as in the proof of \eqref{eq:backf6}. In this proof one uses Condition 5 instead of  \eqref{eq:boundpsipsihat}. One gets that for all $R>0$ we can choose a constant $C_R$ such that if $C_{A3}$ in Condition 3 is large enough with probability tending to one 
$\| \widehat m ^{S} - \widehat m ^{2,S}\|\leq C_R n^{-R} + C C_{A3}^\prime  (1 + C \delta_{2,n})^{C^\prime_{A3} (\log n)^2}  \delta_{1,n}  \leq C_R n^{-R} + C  \delta_{1,n}$. If $R$ is chosen large enough we get the following lemma, see Assumption Condition 4.

    \begin{lemma}
        \label{lem4} If $C_{A3}$ in Condition 3 is large enough it holds that with probability tending to one
	    \begin{eqnarray*}
	        \| \widehat m ^{S} - \widehat m ^{2,S}\|\leq C  \delta_{1,n}.	
	    \end{eqnarray*} 	
	 \end{lemma}
	 We now define  $\widehat m ^{3,S} = \sum_{t \in T_r} \widehat m ^{3,S}_t$ as a minimizer of $$ \sum_{i=1} ^n \left(Y_i -  \sum_{t \in T_r} m_t(X_{i,t})\right)^2$$
	 over all function $m_t: (0,1]^{|t|} \to \mathbb{R}$ that are piecewise constant on the leaves  $\mathcal{I}^S_{t,l}: l=1 ,\dots,L_{t}^S$. Then we have $\widehat \Psi_{t}^{S} \widehat m ^{3,S} +  \widehat{  m} ^{*,S}_t = \widehat m ^{3,S}$ for all $t \in T_r$ which implies that  
	 \begin{eqnarray} \label{eq:seriesm3} \widehat m ^{3,S} &=& \widehat{  m} ^{*,S} _{t_{S}} + \widehat \Psi_{t_S}^{S}  \widehat{  m} ^{*,S} _{t_{S-1}}
 + \widehat \Psi_{t_S}^{S} 
  \widehat \Psi_{t_{S-1}}^{S}
 \widehat{  m} ^{*,S}_{t_{S-2}}
+ ...+  \widehat \Psi_{t_S}^{S}...\widehat \Psi_{t_{S^*+1}}^{S}  \widehat{  m} ^{3,S}_{t_{S^*}}  . \end{eqnarray}  
This shows that 
 \begin{eqnarray*} \widehat m ^{3,S}- \widehat m ^{2,S} &=&  \widehat \Psi_{t_S}^{S}...\widehat \Psi_{t_{S^*+1}}^{S} ( \widehat{  m} ^{3,S}_{t_{S^*}}  -  \widehat{  m} ^{*,S}_{t_{S^*}}). \end{eqnarray*}  
This equation can be used to prove the following result:

\begin{lemma}\label{lem44} If $C_{A3}$ in Condition 3 is large enough it holds that with probability tending to one
	 \begin{eqnarray*}\| \widehat m ^{S} - \widehat m ^{3,S}\| \leq C \delta_{1,n}.	\end{eqnarray*} 	\end{lemma}
	 
	 We now decompose $ \widehat m ^{3,S}$ into a stochastic and a bias term $$ \widehat m ^{3,S} - m^0 = \widehat m ^{A,S}  + \widehat m ^{B,S} - m^0,$$ where 
	 $\widehat m ^{A,S} = \sum_{t \in T_r} \widehat m ^{A,S}_t$ minimizes $$ \sum_{i=1} ^n \left(\varepsilon_i -  \sum_{t \in T_r} m_t(X_{i,t})\right)^2$$
	 over all function $m_t: (0,1]^{|t|} \to \mathbb{R}$ that are piecewise constant on the leaves  $\mathcal{I}^S_{t,l}: l=1 ,...,L_{t}^S$ and $\widehat m ^{B,S} = \sum_{t \in T_r} \widehat m ^{B,S}_t$ minimizes $$ \sum_{i=1} ^n \left(m^0(X_i) -  \sum_{t \in T_r} m_t(X_{i,t})\right)^2$$ over the same class of piecewise constant functions. We see that $\widehat m ^{A,S} $ is the projection of $\varepsilon$ onto an $(S+1)$-dimensional linear subspace of $\mathbb{R}^n$. We conclude that 
	 \begin{eqnarray}
	    E \left [\frac 1 n \sum_{i=1} ^n  \left | \sum_{t \in T_r} \widehat m^{A,S}_t(X_{i,t}) \right |^2 \right]&\leq& (S +1) n^{-1} \sigma^2. \label{eq:boundAterm}
	\end{eqnarray}
	For the study of the bias term define $\bar m(x) = \sum_{t\in T_r} \bar m_t(x_t)$ with 
    $$\bar m_t(x_t) = \frac{1}{n|\mathcal{I}^S_{t,l}|_n} \sum_{i:X_{i,t} \in  \mathcal{I}^S_{t,l}}m^0_t(X_{i,t}) $$
	for $x_t \in \mathcal{I}^S_{t,l}$.
	Now, by definition of $\widehat m ^{B,S} $, we have that 
    $$\frac 1 n \sum_{i=1} ^n  \left | \sum_{t \in T_r} \widehat m^{B,S}_t(X_{i,t}) - m^0(X_i) \right |^2 \leq \frac 1 n \sum_{i=1} ^n  \left | \bar m(X_{i}) - m^0(X_i) \right |^2.$$
	Furthermore, we have by an application of Condition 4 that 
	\begin{eqnarray*} 
	    \frac 1 n \sum_{i=1} ^n  \left | \bar m(X_{i}) - m^0(X_i) \right |^2\leq C \frac{1}{n}\sum_{i=1}^n\sum_{t \in T_r} \sum _{k \in t}  \left(b^S_{t,l_t(X_{i,t}),k}- a^S_{t,l_t(X_{i,t}),k}\right)^2 \leq C \delta_{1,n}^2.
    \end{eqnarray*}
    Using the same arguments as in the proof of Lemma \ref{lem2} one gets the same bound with empirical norm replaced by the $L_2(P)$ norm $\|\cdot\|$. This concludes the proof of the theorem.

  \subsection{Proof of Theorem \ref{theomain2}} \label{prooftheo2}  We now introduce the super index $v$ again which denotes the number of the respective tree . Note that the bound of Lemma \ref{lem44} holds uniformly over $1 \leq v \leq V$. We can decompose  $ \widehat m ^{3,S,v}$ into a stochastic and a bias term $$ \widehat m ^{3,S,v} - m^0 = \widehat m ^{A,S,v}  + \widehat m ^{B,S,v} - m^0$$ and we get from \eqref{eq:boundAterm} that 
    \begin{eqnarray*}
        E \left [\frac 1 n \sum_{i=1} ^n  \left | \sum_{t \in T_r} \frac 1 V \sum_{v=1} ^V  \widehat m^{A,S,v}_t(X_{i,t}) \right |^2 \right]&\leq& (S +1) n^{-1} \sigma^2. 
	\end{eqnarray*}
	For the treatment of the averaged bias term note that for $t \in T_r$
	\begin{eqnarray*}
	    \widehat m^{B,S,v}_t (x_t) = \widehat {\bar m}_t^{B,S,v}(x_t)		 - \sum_{t'\not = t} \int \frac {\widehat p^{S,v} _{t\cup t'} (x_{t\cup t'})}	{\widehat p^{S,v} _{t} (x_{t})} 
		\widehat m^{B,S,v}_{t'} (x_{t'}) \mathrm d x_{t'\backslash t}
	\end{eqnarray*}
	where $ \widehat {\bar m}_t^{B,S,v}(x_t)	 = \frac{1}{n|\mathcal{I}_{t,l}^{S,v} |_n} \sum_{i: X_{i,t} \in \mathcal{I}_{t,l}^{S,v}} m^0 (X_i)$ for $x_t \in \mathcal{I}_{t,l}^{S,v}$. Now by subtracting $m^0_t(x_t)$ we get  
	\begin{eqnarray*}
	    &&\widehat m^{B,S,v}_t (x_t)- m^0_t(x_t) = \widehat {\bar m}_t^{B,S,v}(x_t)		-\bar m^0_t(x_t) \\
		&&\qquad - \sum_{t'\not = t} \int \left (\frac {\widehat p^{S,v} _{t\cup t'} (x_{t\cup t'})}	{\widehat p^{S,v} _{t} (x_{t})} 
		\widehat m^{B,S,v}_{t'} (x_{t'}) - 
		\frac {p_{t\cup t'} (x_{t\cup t'})}	{ p_{t} (x_{t})} 
		m^0_{t'} (x_{t'})\right )\mathrm d x_{t'\backslash t}
	\end{eqnarray*}
	where $\bar m^0_t(x_t) = \int \frac {p(x)}{L_t{(x_t)}} m^0(x) \mathrm d x_{\{1,...,d\}\backslash t}$. This can be rewritten as 
	\begin{eqnarray*}
	    &&\widehat m^{B,S,v}_t (x_t)- m^0_t(x_t) = \widehat {\bar m}_t^{B,S,v}(x_t)		-\bar m^0_t(x_t)  +\Delta_{1,t}^v(x_t)\\
		&&\qquad - \sum_{t'\not = t} \int  \frac {p_{t\cup t'} (x_{t\cup t'})}	{ p_{t} (x_{t})} 
		\left ( \widehat m^{B,S,v}_{t'} (x_{t'}) - 
		m^0_{t'} (x_{t'})\right )\mathrm d x_{t'\backslash t} +\Delta_{2,t}^v(x_t),
	\end{eqnarray*}
	where 
	\begin{eqnarray*}
	    && \Delta_{1,t}^v(x_t) = - \left (\widehat p^{S,v} _{t} (x_{t}) -  p_{t} (x_{t}) \right )  \sum_{t'\not = t} \int  \frac {p_{t\cup t'} (x_{t\cup t'})}	{ p^2_{t} (x_{t})} m^0_{t'} (x_{t'})\mathrm d x_{t'\backslash t}\\
		&& \qquad - \sum_{t'\not = t}  \int  \frac {\widehat p^{S,v} _{t\cup t'} (x_{t\cup t'}) - p_{t\cup t'} (x_{t\cup t'})}	{ p_{t} (x_{t})} 
		m^0_{t'} (x_{t'}) \mathrm d x_{t'\backslash t} ,\\
		&& \Delta_{2,t}^v(x_t) =    \sum_{t'\not = t}  \int \left ( \frac {\widehat p^{S,v} _{t\cup t'} (x_{t\cup t'}) }	{\widehat p^{S,v}_{t} (x_{t})} - \frac { p_{t\cup t'} (x_{t\cup t'})}	{ p_{t} (x_{t})} \right )
		\left ( \widehat m^{B,S,v}_{t'} (x_{t'})- m^0_{t'} (x_{t'}) \right ) \mathrm d x_{t'\backslash t}\\
		&& \qquad -\frac { \left (\widehat p^{S,v} _{t} (x_{t}) -  p_{t} (x_{t}) \right )^2 } { p^2_{t} (x_{t}) \widehat p^{S,v} _{t} (x_{t}) } \sum_{t'\not = t} \int  \widehat p^{S,v} _{t\cup t'} (x_{t\cup t'}) \widehat m^{B,S,v}_{t'} (x_{t'})\mathrm d x_{t'\backslash t}\\
		&& \qquad +\frac { \left (\widehat p^{S,v} _{t} (x_{t}) -  p_{t} (x_{t}) \right ) } { p^2_{t} (x_{t})  } \sum_{t'\not = t} \int  p_{t\cup t'} (x_{t\cup t'}) (\widehat m^{B,S,v}_{t'} (x_{t'})- m^0_{t'} (x_{t'}) ) \mathrm d x_{t'\backslash t}.
	\end{eqnarray*}
	By averaging the integral equations over $v$ we get
	\begin{eqnarray} \label{eq:forestm1}
	    &&\widehat m^{B,S,+}_t (x_t)- m^0_t(x_t) = \widehat {\bar m}_t^{B,S,+}(x_t)		-\bar m^0_t(x_t)  +\Delta_{1,t}(x_t)\\ \nonumber
		&&\qquad - \sum_{t'\not = t} \int  \frac {p_{t\cup t'} (x_{t\cup t'})}	{ p_{t} (x_{t})} 
		\left ( \widehat m^{B,S,+}_{t'} (x_{t'}) - 
		m^0_{t'} (x_{t'})\right )\mathrm d x_{t'\backslash t} +\Delta_{2,t}(x_t),
	\end{eqnarray} 
	where $m^{B,S,+}_{t}= V^{-1} \sum_{v=1}^V m^{B,S,v}_{t}$, $\widehat {\bar m}_t^{B,S,+}= V^{-1} \sum_{v=1}^V \widehat {\bar m}_t^{B,S,v}$, $\Delta_{1,t} = V^{-1} \sum_{v=1}^V\Delta_{1,t}^v $
    and $\Delta_{2,t}= V^{-1} \sum_{v=1}^V \Delta_{2,t}^v$.
    
     We now argue that with probability tending to one
	\begin{eqnarray}
	    &&\label{eq_Delta*1} \| \Delta_{1,t}\|_{1} \leq C \delta_{4,n},\\
		&& \label{eq_Delta*2} \| \Delta_{2,t}\|_1 \leq C (\delta_{2,n} (\delta_{1,n}  + S^{1/2} n^{-1/2})+
        \delta_{3,n}^2),\\
		&& \label{eq_Delta*3}\| \widehat {\bar m}_t^{B,S,+}	-\bar m^0_t\|_{1}  \leq  C (\delta_{4,n} + \delta_{3,n}^2).
	\end{eqnarray}
    Note that with $\Delta_t = \Delta_{1,t} + \Delta_{2,t}+\widehat {\bar m}_t^{B,S,+}	-\bar m^0_t$ we can rewrite \eqref{eq:forestm1} as
    \begin{eqnarray} \label{eq:forestm2}
        \widehat m^{B,S,+}- m^0 = \Delta_t + \Psi_t (\widehat m^{B,S,+}- m^0).
    \end{eqnarray} 
    Order the elements of $T_r$ as $t_1,...,t_{2^r}$. Iterative application of \eqref{eq:forestm2} gives that 
    \begin{eqnarray} \label{eq:forestm3}
        \widehat m^{B,S,+}- m^0 = \Delta_+ + \Psi_+ (\widehat m^{B,S,+}- m^0), 
    \end{eqnarray}
    where $\Delta_+ = \Delta_{t_1} + \Psi_{t_1}\Delta_{t_2} + ... + \Psi_{t_1} ... \Psi_{t_{2 ^r-1}}\Delta_{t_{2 ^r}}$ and $\Psi_+ = \Psi_{t_1} ... \Psi_{t_{2 ^r}}$.
    Because $\Psi_+$ is a complete operator we get from \eqref{eq:Kbound} that $\| \Psi_+ \mu\| \leq \gamma \| \mu \|$ for $\mu \in \mathcal H_{add}$ with $\gamma <1$.
    Furthermore, one can easily verify $\| \Psi_+ \mu\| \leq C \| \mu \|_1$ for $\mu \in \mathcal H_{add}$ with some constant $C >0$. With the help of \eqref{eq_Delta*1}--\eqref{eq_Delta*3} this shows $\|\Delta_+\|_1 \leq C \delta_n$ and $\|\Psi_+ \Delta_+\| \leq C \delta_n$ with probability tending to one where $\delta_n= \delta_{2,n} (\delta_{1,n}  + S^{1/2} n^{-1/2})+
    \delta_{3,n}^2 + \delta_{4,n}$. 
    From \eqref{eq:forestm3} we get $\widehat m^{B,S,+}- m^0 = \Delta_+ + \Delta_{++}$ with $\Delta_{++} = \sum_{k=1}^\infty \Psi^k_+ \Delta_+$.
    With probability tending to one, it holds  $\|\Delta_{++}\| \leq C \delta_n$ which implies $\|\Delta_{++}\|_1 \leq C \delta_n$. We conclude $\|\widehat m^{B,S,+}- m^0\|_1 \leq C \delta_n$ with probability tending to one. This shows the statement of Theorem \ref{theomain2}. It remains to verify \eqref{eq_Delta*1}--\eqref{eq_Delta*3}. The first claim follows directly from Condition 7. 
    For the proof of \eqref{eq_Delta*2} one makes use of the bounds for 
    $\widehat m^{B,S,v}_{t'}  - m^0_{t'} $, $\widehat p^{S,v} _{t\cup t'} - p_{t\cup t'}$ and $\widehat p^{S,v} _{t} -p_t$ which we have shown during the proof of Theorem \ref{theomain1} and which carry over to the averaged values. For the proof of \eqref{eq_Delta*3} note that 
    \begin{eqnarray*}	
        && \widehat {\bar m}_t^{B,S,+}(x_t)	-\bar m^0_t (x_t)  = \Delta_{3,t}(x_t) + \Delta_{4,t}(x_t)
	\end{eqnarray*}
	holds with 
	\begin{eqnarray*}
	    && \Delta_{3,t}(x_t) =  \frac {\widehat p^{S,+} _{t} (x_{t}) -  p_{t} (x_{t}) } {p_{t} (x_{t}) }  \sum_{t'\not = t} \int  p_{t\cup t'} (x_{t\cup t'})	 m^0_{t'} (x_{t'})\mathrm d x_{t'\backslash t}\\
		&& \qquad + \sum_{t'\not = t}  \int  \frac {\widehat p^{S,+} _{t\cup t'} (x_{t\cup t'}) - p_{t\cup t'} (x_{t\cup t'})}	{ p_{t} (x_{t})} 
		m^0_{t'} (x_{t'}) \mathrm d x_{t'\backslash t} ,\\
		&& \Delta_{4,t}(x_t) = -  V^{-1} \sum_{v=1} ^V \left (\widehat p^{S,v} _{t} (x_{t}) -  p_{t} (x_{t}) \right ) \\
		&& \qquad \qquad \times  \sum_{t'\not = t} \int  \frac {\widehat p^{S,v} _{t\cup t'} (x_{t\cup t'}) -p_{t\cup t'} (x_{t\cup t'})}	{ p^2_{t} (x_{t})}   m^0_{t'} (x_{t'})  \mathrm d x_{t'\backslash t}\\
		&& \qquad +  V^{-1} \sum_{v=1} ^V \frac { \left (\widehat p^{S,v} _{t} (x_{t}) -  p_{t} (x_{t}) \right )^2 } { p^2_{t} (x_{t}) \widehat p^{S,v} _{t} (x_{t}) } \sum_{t'\not = t} \int  \widehat p^{S,v} _{t\cup t'} (x_{t\cup t'}) m^0_{t'} (x_{t'}) \mathrm d x_{t'\backslash t}.
	\end{eqnarray*}
	Using similar arguments as above, one can easily verify $\| \Delta_{3,t}\| _{t,*}\leq C \delta_{4,n}$ and 
	$ \| \Delta_{4,t}\| \leq C   \delta_{3,n}^2  $ with probability tending to one. This concludes the proof of the theorem.


	
	\bibliographystyle{chicago}
	\bibliography{mybib}
\newpage
 \appendix




\section{The Random Planted Forest Algorithm}\label{Section: The Random Planted Forest Algorithm}

    In this section we add some technical details which improve the understanding of the rpf algorithm. We shortly discuss the parameters introduced and explain why choices were made when designing the algorithm. 
    We then give a short overview over the algorithm in the additive case, i.e. if $\texttt{max\_interaction}=1$. The simplification is easier to understand and of particular interest in our simulation study in Section \ref{simulations}. Next, we include the discussion of an identification constraint for the functional decomposition, which is important for plotting the components.
	
\subsection{$\texttt{max\_interaction}$}
	
	The rpf estimator satisfies BI($\texttt{max\_interaction}$). This is one of the main benefits of rpf. Additionally, $\texttt{max\_interaction}$ can be seen as a tuning parameter of the algorithm. Our simulations show that its value does not have much impact if it is high enough, while there are benefits if one specifies the model precisely. Note that by setting $\texttt{max\_interaction}$ to a high value, one looses the benefit of interpretability. 
	
\subsection{Leaves}
	
	It is important that some leaves may be split multiple times. Consider a naive random forest estimator with the additional condition that leaves with type $|t| = \texttt{max\_interaction}$ are only splittable with respect to dimensions $k \in t$. If we set $\texttt{max\_interaction}=1$, the naive estimator would only ever depend on 1 coordinate. Thus the root leaf must be kept in the algorithm. Similar but slightly more involved examples show that one should not delete leaves when they are split with respect to a new dimension. An other idea may be to never delete any leaf when splitting. We found that this considerably increases computational cost while not decreasing the performance of the algorithm.
	
	Furthermore, in the random forest algorithm, the order in which one splits leaves does not make a difference. However, as discussed above, it makes a difference for rpf. It seems most natural to decide on which leaf to split by optimizing via \eqref{equation2}. We note that this procedure increases the computational cost.
	
\subsection{$\texttt{nsplits}$}
	
	The parameter $\texttt{nsplits}$ determines the number of iterations in a planted tree. In random forests other termination criteria are used such as bounding the tree depth from above or the number of data points in a leaf from below. These criteria do not terminate rpf since for example the root is always splittable in both cases.
	
\subsection{$\texttt{t\_try}$}

    In random forests one limits the number of coordinates considered in each iteration step using a parameter $\texttt{m\_try}$. The idea is to reduce the variance by forcing the trees to differ more substantially. Similarly, we limit the number of combinations of leaf types and coordinates considered in each iteration step.
	The number of combinations is defined as a proportion $\texttt{t\_try}$ of the number of viable combinations $|B|$. Thus the number of combinations considered increases as the algorithm unfolds. We assessed many different similar mechanisms in order to restrict the number of combinations used in an iteration step.  First of all, we found that using a random subset of the viable combinations instead of simply restricting the coordinates for splitting is far superior. Roughly speaking, the reason is that the algorithm may lock itself in a tree if all leaves are splitable in each step. The question remains how to quantify the amount of combinations used. Selecting a constant number of viable combinations has the disadvantage of either allowing all combinations in the beginning or having essentially random splits towards the end of the algorithm. Thus the number of combinations considered must depend on the number of viable combinations. While other functional connections between the number of viable and the number of considered combinations are possible, choosing a proportional connection seems natural.
    
\subsection{$\texttt{split\_try}$} 
    
    In contrast to $\texttt{t\_try}$, the parameter $\texttt{split\_try}$ implies an almost constant number of considered split points in each iteration. The $\texttt{split\_try}$ split options are selected uniformly at random with replacement. In Section \ref{simulations}, we find that the optimal value for $\texttt{split\_try}$ is usually small compared to the data size. In our experience, using a proportion of the split points available either leads to totally random splits for small leaves or essentially allows all splits for large leaves. The reason is that for large leaves, if a small number of data points are added or removed from the leaf, the estimation is basically the same. The version we use here yielded the best results. The choice $\texttt{t\_try}<1$ as well as only using $\texttt{split\_try}$ split points is done in order to reduce the variance of the estimator by reducing correlation between trees. It also reduces computational cost. While it is not obvious which of the two parameters, $\texttt{split\_try}$ or $\texttt{t\_try}$, should be lowered in order to reduce variance, our results significantly improved as soon as we introduced the mechanisms regarding the parameters.

\subsection{Driving Parameters}
    
    Although the presence of the mechanisms involving $\texttt{split\_try}$ and $\texttt{t\_try}$ improve the results of rpf, the exact value they take on is not as relevant. Rather, we consider them to be fine-tuning parameters. Similarly, the number of trees $\texttt{ntrees}$ in a forest does not have much impact as long as $\texttt{ntrees}$ is large enough; in our simulation study, $\texttt{ntrees}=50$ seemed satisfactory. The main driving parameter for the estimation quality of rpf is the number of iteration steps $\texttt{nsplits}$. While the estimation is quite stable under small changes of $\texttt{nsplits}$, strongly lowering $\texttt{nsplits}$ results in a bias, while vastly increasing the parameter leads to overfitting.
	
\subsection{Random Planted Forests for BI($1$)}\label{Additive Random Trees}

	In this section we explain the rpf algorithm in the additive case, i.e. if we set $\texttt{max\_interaction}=1$. Recall that the resulting estimator satisfies BI($1$). The algorithm then simplifies in the following way. All non-root leaves grown during the algorithm have leaf type $\{k\}$ for some $k\in\{1,\dots,d\}$. The set $B$ introduced in Subsection \ref{From a tree  to a Forest} reduces to
	\begin{align*}
	    B=\big\{(\{k\},k)\ \big|\ k\in \{1,\dots,d\}\}.
	\end{align*}
	In particular, it does not depend on the current state of the  of planted tree. Noting that $|B|=d$, the value $\texttt{m\_try}:=\lceil \texttt{t\_try}\cdot d\rceil$ is constant throughout the algorithm. Thus the parameter $\texttt{t\_try}$ or equivalently $\texttt{m\_try}$ act exactly the same as the parameter $\texttt{m\_try}$ in Breimans implementation of random forests. In this case, rpf differs from extremely random forest mainly due to the fact that the root $\mathcal{I}_{\emptyset,1}$ is never deleted during the algorithm and all leaves are constructed using only one coordinate. 
	
	
\subsection{Identification Constraint}\label{Identification Constraint}
	
	If $r\in\{1,2\}$, the components can be visualized easily. In order to reasonably compare plots, we need a condition that ensures uniqueness of the functional decomposition. Note that in contrast to the problem considered by \cite{apley2020visualizing}, the constraint itself is of secondary importance. The reason is that we do not aim to approximate a multivariate estimator by 
	(\ref{anova1}).  The rpf estimator is already in the form of (\ref{anova1}). Hence, the constraint does not have an effect on the quality of the estimator. One possible constraint family is that for every $u\subseteq \{1,\dots,d\}$ and $k\in u$,
	\begin{align}\label{constraint}
	\int m_{u}\left(x_{u}\right) \int w(x) \mathrm dx_{-u} \ \mathrm d x_{k}=0,
	\end{align} 
	for some weight function $w$. Another option is assuming that for every $u\subseteq \{1,\dots,d\}$,
\begin{align}\label{constraint2}
\sum_{v \cap u \neq \emptyset} \int \hat m_v(x_v) \hat p_u(x_u)\mathrm dx_u=0,
\end{align}
where $\hat p_u$ is some estimator of the density $p_u$ of $X_u$, see \cite{hiabu2022unifying}.
Observe that the constraints are not an additional assumption on the overall function $m$. The constraint ensures that the functional decomposition is unique for non-degenerated cases. One possibility for the weight in \eqref{constraint} is $w(x)=\prod_k \widehat p_k(x_k)$. With the burden of extra computational cost, one can choose $w$ as an estimate of the full design density, see also \cite{lengerich2020purifying}. Compared to the previous example, a computationally more efficient constraint has recently been proposed in \cite{apley2020visualizing}. Depending on the viewpoint, different choices for $w$ may be advantageous. Since the precise constraint is not the focus of this paper, we settled for \eqref{constraint} with the simple constraint $w\equiv 1$ which suffices in the sense that it allows us to compare plots of different estimators.

	Note that while we obtain an estimator $\widehat{m}_t$ for every component in \eqref{anova1}, in general, these estimators do not satisfy the constraint (\ref{constraint}). Thus in order to obtain suitable estimators for the components, we must normalize the estimated components without changing the overall estimator. This can be achieved by using an algorithm similar to the purification algorithm given in \cite{lengerich2020purifying}.

\section{Example for a Theoretical Random Planted Forest Algorithm}\label{Example for a Theoretical Random Planted Forest Algorithm}

We now describe in more detail the iteration steps of a tree  algorithm that motivates the setting in Section \ref{Theoretical properties in the additive case}. As initialisation we set $\mathcal{I}^0_{t,1}= [0,1]^d$, $L_{t,0}=1$ and $\widehat m^{0}_{t } \equiv 0$ for $t \in T_r$. In iteration steps $s=1,...,S$ partitions $\mathcal{I}^s_{t,l}=\prod_{k \in t} (a_{t,l,k}^s, b^s_{t,l,k}] \times \prod_{k \not \in t} (0,1]$ of $[0,1]^d$ with $l=1,...,L^s_{t}$ are updated by splitting one of the leaves $\mathcal{I}^s_{t,l}$ for one $t \in T_r$, one $l \in \{ 1, ...,L^s_{t}\}$ along one coordinate $k$. Here in abuse of notation,  we write $\prod_{k \in t} \cdot \times \prod_{k \not \in t} \cdot$ for the set of tuples with coordinates ordered according to the value of $k$ and not according to the appearance in the product sign subindices. We now describe  step $s$ where the leaves $\mathcal{I}^{s-1}_{t,l}$ are updated. At step $s$ one chooses a $t_s \in T_r$, an $l_s \in \{1,...,L_{t_s}^{s-1}\}$, a $k_s \in t_s$ and a splitting value $b_{t_s,l_s,k_s}^s\in (a_{t_s, l_s,k_s}^{s-1}, b_{t_s, l_s,k_s}^{s-1})$. The values are chosen by some random procedure that satisfies the conditions in Section \ref{Theoretical properties in the additive case}. Using these values one splits $(a_{t_s,l_s,k_s}^{s-1}, b_{t_s,l_s,k_s}^{s-1}]$ into $(a_{t_s,l_s,k_s}^{s}, b_{t_s,l_s,k_s}^{s}]$ and $(a^{s}_{t_s,L_{t_s}^s,k_s}, b^{s}_{t_s, L_{t_s}^s,k_s}]$, where $a_{t_s,l_s,k_s}^{s}= a_{t_s,l_s,k_s}^{s-1}$, $a^s_{t_s, L_{t_s}^s,k_s}= 	b_{t_s,l_s,k_s}^{s}$,  $b^s_{t_s,L_{t_s}^s,k_s}= 	b_{t_s,l_s,k_s}^{s-1}$ and $L_{t_s}^s=L_{t_s}^{s-1}+1$. For $(l, k) \not = (l_s, k_s)$ we set $(a_{t_s,l, k}^{s}, b_{t_s,l, k}^{s}]=(a_{t_s,l, k}^{s-1}, b_{t_s,l, k}^{s-1}]$. Then we update the leaf $\mathcal{I}^s_{t_s,l}$ for $l=l_s$ and $l = L_{t_s}^s$ by defining $\mathcal{I}^s_{t_s,l} =\mathcal{I}^{\text{marg},s}_{t_s,l}  \times \prod_{k \not \in t_s} (0,1]$ with $\mathcal{I}^{\text{marg},s}_{t_s,l} =\prod_{k \in t_s} (a_{t_s,k,l}^s, b^s_{t_s,k,l}]$. All other leaves are taken over identically from the last step. Finally, for the chosen tree $t_s$,  the values $\widehat m_{t_s}^s$ are updated by averaging  residuals over the intervals $\mathcal{I}^{s}_{t_s,l}$ for $l= 1 , ..., L_{t_s}^s$. Then for $x \in \mathcal{I}^{s}_{t_s,l} $ with $l \in \{1,...,L_{t_s}^s\}$ the estimator can be written as
	\begin{eqnarray*} 
	    \widehat m_{t_s} ^s (x ) &=& \frac 1 {n|\mathcal{I}^{s}_{t_s,l}|_n} \sum_{i: X_{i} \in \mathcal{I}^{s}_{t_s,l}} \left(Y_i -  \sum _{t \in T_r, t \not = t_s} \widehat m^{s-1}_{t } (X_{i})\right).	
	\end{eqnarray*} 
	Because $\widehat m^{s-1}_{t}$ is constant on the set $\mathcal{I}^{s}_{t,l} $ we can rewrite the formula with $\widehat m^{s-1}_{t, l'}$ equal to $\widehat m^{s-1}_{t } (u)$ for $u \in \mathcal{I}^{s}_{t,l'} $ as follows
	\begin{eqnarray*} 
	    \widehat m_{t_s} ^s (x_{t_s} ) &=& \widehat{  m}_{t_s} ^{*,s} (x_{t_s} )  - \frac 1 {|\mathcal{I}^{s}_{t_{s},l}|_n} \sum _{t \in T_r, t \not = t_s} \sum_{l'=1} ^{L_{t,s}} |\mathcal{I}^{s}_{t_s,l} \cap \mathcal{I}^{s}_{t,l'} |_n\ \widehat m^{s-1}_{t, l'}.
	\end{eqnarray*}
	One can easily check that in the notation of Section \ref{Theoretical properties in the additive case} the estimator is in the form \eqref{eq:backf}.
	
	When one compares the estimator described here and in Section \ref{Theoretical properties in the additive case}
 there are three further differences from the rpf algorithm as decribed in the first sections of the main part of the paper.  First now we construct trees based on the full sample and not on bootstrap samples. As we have seen in the results of Section \ref{Theoretical properties in the additive case} bootstrap is not necessary, neither for rate optimality for $r \leq 2$ nor rate improvement by averaging the trees. Our results can be generalized to versions that make use of bootstrap. Second for each "leaf type" $t\in T_r$ we grow one tree with an own root and  do not allow for more trees of the same leaf type that grow in parallel. This modification is made to simplify mathematical theory. Third in each iteration step $s$ we update the estimator $\widehat m^s_{t_s} (x) $ on all leaves $\mathcal{I}^s_{t_s,l}$ ($l=1,..., L_{t_s}^s$) of "leaf type" $t_s$. In our practical implementation of rpf we only did this for the splitted leaf, i.e. for $l=l_s$ and $l=L_{t_s}^s$. From simulations we concluded that the third change is not severe.

\section{Further Simulation Results}\label{Section:Further Simulation Results}
	
	In this section we provide further results from our simulation study omitted in the main part of this paper.
	The results are given in Tables \ref{parameters}--\ref{lastTable} and further confirm the discussion of Section 5.

    \begin{table}[ht]
		\centering
		\caption{For each method, we ran 40 simulations  to find the optimal parameter combinations from the parameter range below, measured via sample mean squared error: $n^{-1}\sum_i (m(X_i)-\widehat m(X_i))^2$. These simulations where conducted independently from the final simulations using training data. The names of the parameters are drawn from their functional definitions in their respective R-packages. 
		} \label{parameters}
		\begin{tabular}{|ll|}
			\hline
			Method  & Parameter range\\ 
			\hline
			\hline
			xgboost &  $\texttt{max.depth} = 1$ (if \texttt{depth = 1}), \\& \qquad \qquad   \quad = 2,3,4,  \\
			& $\texttt{eta}$ = $0.005, 0.01, 0.02, 0.04, 0.08, 0.16, 0.32,$  \\
			& $\texttt{nrounds} =100, 300, 600, 1000 , 3000, 5000, 7000.$\\
			rpf & 
			$\texttt{max\_interaction} = 1$ (if $\texttt{max\_interaction = 1}$),  \\& \qquad \qquad   \qquad  \qquad = 2 (if $\texttt{max\_interaction=2}$),  \\& \qquad \qquad  \qquad   \qquad  = $1,2,3,4,30$, \\
			&$\texttt{t\_try}$  = $0.25,0.5,0.75,$\\
			&\texttt{nsplits} = $10,15,20,25,30,40,50,60,80,100, 120, 200$ , \\ 
			& $\texttt{split\_try}$ = $ 2,5,10,20.$ \\
			rf &
			$\texttt{m\_try}$ = $\floor{d/4}, \floor{d/2}, \floor{3d/4},\floor{7d/8},\floor{d},$ \\
			& $\texttt{min.node.size} = 5,$ \\
			& $\texttt{ntrees} = 500$,\\
             &  $\texttt{replace}=\text{TRUE,FALSE}$.\\
			sbf&
			$\texttt{bandwidth} = 0.1,0.2,0.3.$ \\
			gam&
			$\texttt{select}$ =TRUE, $\texttt{method}$ = 'REM' (in sparse models),   \\
			&  default settings (in dense models). \\
			BART & $\texttt{power} (\beta) = 1,2,3$,\\
            & \texttt{ntree} = 50,100,150,200,250,300,\\
            & sparsity parameter = 0.6,0.75,0.9.\\
            MARS & $\texttt{degree} = 1,2,3,4,5,6,7,8,9,10$,\\
            & $\texttt{penalty} = 1,2,3,4,5,6,7,8,9,10$.\\
			
			\hline
		\end{tabular}
	\end{table}
 
    \begin{table}[ht]
		\centering
		\caption{Model 2: Hierarchical Interaction Sparse Smooth Model. We report the average MSE from 100 simulations. Standard deviations are provided in brackets.} 
		\begin{tabular}{llllc}
		\hline
			Method  & dim=4 & dim=10 & dim=30 & BI($r$) $r=0/1/2$ \\ 
            \hline
            xgboost (\texttt {depth=1}) & 2.435 (0.157) & 2.542 (0.15) & 2.587 (0.152) & 1  \\ 
            xgboost  & 0.374 (0.035) & 0.481 (0.064) & 0.557 (0.089) & $\times$ \\ 
            xgboost-CV & 0.393 (0.051) & 0.499 (0.058) & 0.563 (0.089) & $\times$ \\ 
            rpf (\texttt {max\_interaction=1}) & 2.36 (0.165) & 2.43 (0.17) & 2.404 (0.145) & 1  \\ 
            rpf (\texttt {max\_interaction=2}) & 0.248 (0.038) & 0.327 (0.045) & 0.408 (0.07) & 2  \\ 
            rpf  & 0.263 (0.034) & 0.357 (0.044) & 0.452 (0.076) & $\times$ \\ 
            rpf-CV & 0.277 (0.039) & 0.366 (0.051) & 0.463 (0.083) & $\times$ \\ 
            rf  & 0.432 (0.039) & 0.575 (0.061) & 0.671 (0.08) & $\times$ \\ 
            sbf & 2.298 (0.168) & 2.507 (0.181) & 3.163 (0.207) & 1 \\ 
            gam  & 2.242 (0.172) & 2.311 (0.159) & 2.277 (0.185) & 1 \\ 
            BART& 0.214 (0.03) & 0.223 (0.04) & 0.252 (0.037) & $\times$ \\ 
            BART-CV  & 0.242 (0.043) & 0.276 (0.053) & 0.315 (0.047) & $\times$ \\ 
            MARS & 0.355 (0.089) & 0.282 (0.038) & 0.414 (0.126) & $\times$ \\ 
            1-NN  & 2.068 (0.156) & 5.988 (0.624) & 11.059 (0.676) & $\times$ \\ 
            average & 8.366 (0.43) & 8.086 (0.246) & 8.207 (0.496) & 0  \\ 
            \hline
		\end{tabular}
	\end{table}
 
    \begin{table}[ht]
		\centering
		\caption{Model 3: Pure Interaction Sparse Smooth Model. We report the average MSE from 100 simulations. Standard deviations are provided in brackets.}\label{Pure Interaction-Sparse-Smooth}
		\begin{tabular}{llllc}
		\hline
			Method  & dim=4 & dim=10 & dim=30 & BI($r$) $r=0/1/2$ \\ 
            \hline
            xgboost (\texttt {depth=1}) & 2.176 (0.14) & 2.236 (0.176) & 2.183 (0.136) & 1 \\ 
            xgboost  & 0.417 (0.082) & 0.797 (0.16) & 1.381 (0.234) & $\times$ \\ 
            xgboost-CV  & 0.443 (0.078) & 0.872 (0.136) & 1.497 (0.326) & $\times$ \\ 
            rpf (\texttt {max\_interaction=1}) & 2.172 (0.133) & 2.236 (0.164) & 2.199 (0.145) & $1$ \\ 
            rpf(\texttt {max\_interaction=2})& 0.416 (0.082) & 1.289 (0.224) & 1.822 (0.208) & $2$ \\ 
            rpf  & 0.219 (0.035) & 0.556 (0.143) & 1.186 (0.236) & $\times$ \\ 
            rpf-CV & 0.233 (0.033) & 0.603 (0.163) & 1.313 (0.253) & $\times$ \\ 
            rf  & 0.304 (0.047) & 0.744 (0.305) & 1.295 (0.317) & $\times$ \\ 
            sbf  & 2.249 (0.159) & 2.473 (0.181) & 3.133 (0.22) & 1 \\ 
            gam & 2.161 (0.13) & 2.222 (0.172) & 2.209 (0.168) & 1 \\ 
            BART  & 0.168 (0.022) & 0.172 (0.032) & 0.202 (0.021) & $\times$ \\ 
            BART-CV & 0.192 (0.03) & 0.199 (0.039) & 0.223 (0.025) & $\times$\\ 
            MARS & 0.245 (0.088) & 0.831 (0.728) & 0.429 (0.403) & $\times$\\ 
            1-NN  & 1.323 (0.117) & 2.642 (0.317) & 4.173 (0.413) &$\times$\\ 
            average & 2.187 (0.125) & 2.226 (0.174) & 2.177 (0.146) & 0  \\ 
            \hline
		\end{tabular}
	\end{table}
	
	\begin{table}[ht]
		\centering
		\caption{Model 5: Hierarchical Interaction Sparse  Jump Model. We report the average MSE from 100 simulations. Standard deviations are provided in brackets.} 
		\begin{tabular}{llllc}
				\hline
			Method  & dim=4 & dim=10 & dim=30 & BI($r$) $r=0/1/2$ \\ 
		    \hline
            xgboost(\texttt {depth=1})& 2.974 (0.112) & 3.046 (0.12) & 3.098 (0.223) & 1 \\ 
            xgboost  & 1.02 (0.152) & 1.28 (0.16) & 1.418 (0.156) & $\times$\\ 
            xgboost-CV  & 1.049 (0.125) & 1.279 (0.157) & 1.475 (0.185) &$\times$ \\ 
            rpf (\texttt {max\_interaction=1}) & 2.941 (0.117) & 2.942 (0.123) & 2.913 (0.197) & 1\\ 
            rpf  (\texttt {max\_interaction=2})& 0.767 (0.096) & 1.082 (0.139) & 1.34 (0.132) & 2 \\ 
            rpf & 0.745 (0.089) & 1.093 (0.142) & 1.307 (0.113) & $\times$ \\ 
            rpf-CV  & 0.769 (0.101) & 1.167 (0.152) & 1.404 (0.14) & $\times$ \\ 
            rf &  0.914 (0.091) & 1.237 (0.121) & 1.415 (0.152) & $\times$ \\ 
            sbf & 2.791 (0.098) & 2.926 (0.12) & 3.756 (0.284) & 1 \\ 
            gam  & 2.782 (0.085) & 2.728 (0.105) & 2.793 (0.208) & 1 \\ 
            BART & 0.611 (0.078) & 0.644 (0.106) & 0.67 (0.094) & $\times$ \\ 
            BART-CV & 0.661 (0.111) & 0.772 (0.173) & 0.791 (0.133) & $\times$ \\ 
            MARS  & 2.306 (0.17) & 2.325 (0.145) & 3.374 (2.716) & $\times$ \\ 
            1-NN & 4.559 (0.409) & 8.883 (0.692) & 13.434 (0.674) & $\times$ \\ 
            average & 8.721 (0.334) & 8.449 (0.229) & 8.638 (0.412) & 0 \\ 
            \hline
		\end{tabular}
	\end{table}
	
	\begin{table}[ht]
		\centering
		\caption{Model 6: Pure Interaction Sparse Jump Model. We report the average MSE from 100 simulations. Standard deviations are provided in brackets.} 
		\begin{tabular}{llllc}
			\hline
			Method & dim=4 & dim=10 & dim=30 & BI($r$) $r=0/1/2$ \\ 
            \hline
            xgboost (\texttt {depth=1}) & 2.662 (0.078) & 2.616 (0.105) & 2.565 (0.153) & 1 \\ 
            xgboost & 1.034 (0.177) & 1.723 (0.178) & 2.337 (0.378) & $\times$\\ 
            xgboost-CV & 1.196 (0.371) & 2.056 (0.3) & 2.481 (0.385) & $\times$ \\ 
            rpf (\texttt {max\_interaction=1})& 2.682 (0.076) & 2.653 (0.103) & 2.601 (0.155) & 1 \\ 
            rpf (\texttt {max\_interaction=2}) & 1.252 (0.164) & 2.268 (0.13) & 2.534 (0.175) & 2 \\ 
            rpf  & 0.834 (0.121) & 1.729 (0.156) & 2.337 (0.284) & $\times$\\ 
            rpf-CV& 0.886 (0.142) & 1.939 (0.2) & 2.438 (0.246) & $\times$ \\ 
            rf  & 0.805 (0.172) & 1.696 (0.168) & 2.276 (0.306) & $\times$ \\ 
            sbf  & 2.757 (0.094) & 2.893 (0.128) & 3.705 (0.282) & 1 \\ 
            gam  & 2.645 (0.096) & 2.617 (0.095) & 2.674 (0.165) & 1 \\ 
            BART  & 0.583 (0.074) & 0.632 (0.124) & 0.798 (0.29) & $\times$ \\ 
            BART-CV & 0.608 (0.106) & 0.73 (0.184) & 1.16 (0.655) & $\times$ \\ 
            MARS  & 2.324 (0.14) & 2.549 (0.296) & 2.522 (0.291) & $\times$ \\ 
            1-NN  & 3.769 (0.323) & 5.459 (0.419) & 6.247 (0.434) & $\times$ \\ 
            average  & 2.637 (0.092) & 2.59 (0.106) & 2.55 (0.14) & 0 \\ 
            \hline
		\end{tabular}
	\end{table}
	
	\begin{table}[ht]
		\centering
		\caption{Model 7: Additive Dense Smooth Model. We report the average MSE from 100 simulations. Standard deviations are provided in brackets.}  
		\label{tab:4}
		\begin{tabular}{lllc}
			\hline
			Method  & dim=4 & dim=10 & BI($r$) $r=0/1/2$ \\ 
            \hline
            xgboost (\texttt {depth=1}) & 0.2 (0.035) & 0.662 (0.059) & 1 \\ 
            xgboost  & 0.273 (0.028) & 1.233 (0.127) & $\times$ \\ 
            xgboost-CV & 0.209 (0.043) & 0.673 (0.06) & $\times$ \\ 
            rpf (\texttt {max\_interaction=1})& 0.162 (0.025) & 0.578 (0.068) & 1 \\ 
            rpf  (\texttt {max\_interaction=2}) & 0.191 (0.017) & 0.798 (0.097) & 2 \\ 
            rpf  & 0.222 (0.019) & 1.052 (0.115) & $\times$\\ 
            rpf-CV & 0.178 (0.03) & 0.6 (0.072) & $\times$ \\ 
            rf  & 0.567 (0.044) & 10.527 (0.772) & $\times$\\ 
            sbf  & 0.071 (0.021) & 0.183 (0.026) & 1  \\ 
            gam & 0.055 (0.012) & 0.171 (0.045) & 1 \\ 
            BART  & 0.155 (0.023) & 0.438 (0.053) & $\times$ \\ 
            BART-CV  & 0.165 (0.032) & 0.465 (0.094) & $\times$ \\ 
            MARS & 0.166 (0.035) & 4.4 (0.36) & $\times$ \\ 
            1-NN  & 2.05 (0.108) & 11.634 (0.702) & $\times$\\ 
            average  & 7.71 (0.381) & 18.986 (1.391) & 0 \\ 
            \hline
		\end{tabular}
	\end{table}

	\begin{table}[ht]
		\centering
		\caption{Model 8: Hierarchical Interaction Dense Smooth Model. We report the average MSE from 100 simulations. Standard deviations are provided in brackets.} \label{tab:5} 
		\begin{tabular}{lllc}
			\hline
			Method & dim=4 & dim=10 & BI($r$) $r=0/1/2$ \\ 
            \hline
            xgboost (\texttt {depth=1}) & 3.509 (0.266) & 10.108 (0.425) & 1 \\ 
            xgboost  & 0.645 (0.053) & 2.895 (0.271) & $\times$ \\ 
            xgboost-CV  & 0.678 (0.042) & 3.013 (0.338) & $\times$ \\ 
            rpf (\texttt {max\_interaction=1})& 3.408 (0.237) & 9.717 (0.378) & 1 \\ 
            rpf (\texttt {max\_interaction=2})& 0.414 (0.047) & 3.643 (0.349) & 2 \\ 
            rpf  & 0.385 (0.034) & 3.357 (0.372) & $\times$ \\ 
            rpf-CV  & 0.413 (0.033) & 3.665 (0.467) & $\times$ \\ 
            rf  & 0.77 (0.034) & 12.265 (1.447) & $\times$ \\ 
            sbf  & 3.42 (0.208) & 9.215 (0.419) & 1 \\ 
            gam& 3.258 (0.227) & 9.212 (0.483) & 1 \\ 
            BART  & 0.34 (0.04) & 1.889 (0.324) & $\times$\\ 
            BART-CV & 0.354 (0.059) & 2.133 (0.363) & $\times$\\ 
            MARS & 0.624 (0.114) & 10.885 (0.635) & $\times$\\ 
            1-NN & 2.516 (0.141) & 17.728 (1.215) & $\times$\\ 
            average & 10.696 (0.621) & 26.502 (1.892) & 0 \\ 
            \hline
		\end{tabular}
	\end{table}
	
%
	
	\begin{table}[ht]
		\centering
		\caption{Model 9: Pure Interaction Dense Smooth Model. We report the average MSE from 100 simulations. Standard deviations are provided in brackets.}  
		\begin{tabular}{lllc}
			\hline
			Method & dim=4 & dim=10 & BI($r$) $r=0/1/2$ \\ 
	        \hline
            xgboost (\texttt {depth=1})& 3.108 (0.197) & 8.091 (0.359) & 1 \\ 
            xgboost  & 0.596 (0.063) & 3.888 (0.411) & $\times$ \\ 
            xgboost-CV  & 0.684 (0.069) & 3.974 (0.508) & $\times$ \\ 
            rpf (\texttt {max\_interaction=1}) & 3.119 (0.209) & 8.156 (0.366) & 1 \\ 
            rpf (\texttt {max\_interaction=2}) & 0.712 (0.101) & 5.944 (0.324) & 2 \\ 
            rpf & 0.38 (0.049) & 4.747 (0.329) & $\times$ \\ 
            rpf-CV & 0.395 (0.055) & 4.789 (0.335) & $\times$ \\ 
            rf  & 0.657 (0.074) & 5.784 (0.409) & $\times$ \\ 
            sbf  & 3.385 (0.183) & 9.177 (0.479) & 1 \\ 
            gam & 3.109 (0.216) & 8.183 (0.389) & 1\\ 
            BART & 0.266 (0.034) & 1.425 (0.183) & $\times$ \\ 
            BART-CV & 0.299 (0.054) & 1.738 (0.254) & $\times$ \\ 
            MARS& 0.618 (0.552) & 6.257 (0.824) & $\times$ \\ 
            1-NN  & 1.482 (0.126) & 7.358 (0.514) & $\times$ \\ 
            average & 3.156 (0.221) & 8.109 (0.363) & 0 \\ 
            \hline
		\end{tabular}
	\end{table}

%
%
	
	\begin{table}[ht]
		\centering
		\caption{Model 10: Additive Dense Jump Model.We report the average MSE from 100 simulations. Standard deviations are provided in brackets.} 
		\begin{tabular}{lllc}
			\hline
			Method & dim=4 & dim=10 & BI($r$) $r=0/1/2$\\ 
	        \hline
            xgboost (\texttt {depth=1}) & 0.325 (0.068) & 1.095 (0.106) & 1 \\ 
            xgboost  & 0.376 (0.085) & 1.437 (0.153) & $\times$ \\ 
            xgboost-CV  & 0.36 (0.073) & 1.187 (0.145) & $\times$ \\ 
            rpf (\texttt {max\_interaction=1})& 0.321 (0.047) & 1.273 (0.161) & 1 \\ 
            rpf (\texttt {max\_interaction=2})& 0.402 (0.059) & 2.18 (0.139) & 2 \\ 
            rpf & 0.429 (0.067) & 2.804 (0.192) & $\times$ \\ 
            rpf-CV & 0.326 (0.051) & 1.303 (0.166) & $\times$ \\ 
            rf  & 0.807 (0.104) & 4.051 (0.186) & $\times$ \\ 
            sbf  & 0.588 (0.078) & 1.685 (0.185) & 1 \\ 
            gam  & 0.923 (0.15) & 4.405 (0.379) & 1 \\ 
            BART  & 0.369 (0.079) & 1.164 (0.093) & $\times$ \\ 
            BART-CV & 0.39 (0.076) & 1.436 (0.253) & $\times$ \\ 
            MARS  & 1.749 (0.151) & 5.424 (0.304) & $\times$ \\ 
            1-NN  & 3.822 (0.296) & 11.278 (1.097) & $\times$ \\ 
            average  & 2.5 (0.116) & 6.332 (0.41) & 0 \\ 
            \hline
		\end{tabular}
	\end{table}


	\begin{table}[ht]
		\centering
		\caption{Model 11: Hierarchical Interaction Dense Jump Model. We report the average MSE from 100 simulations. Standard deviations are provided in brackets.}  
		\begin{tabular}{lllc}
			\hline
			Method  & dim=4 & dim=10 & BI($r$) $r=0/1/2$ \\ 
            \hline
            xgboost(\texttt {depth=1}) & 4.19 (0.238) & 13.112 (0.83) & 1  \\ 
            xgboost & 1.666 (0.159) & 9.327 (0.594) & $\times$ \\ 
            xgboost-CV  & 1.87 (0.312) & 9.407 (0.653) & $\times$ \\ 
            rpf(\texttt {max\_interaction=1}) & 4.19 (0.253) & 12.997 (0.831) & 1 \\ 
            rpf (\texttt {max\_interaction=2}) & 1.43 (0.205) & 9.238 (0.648) & 2 \\ 
            rpf & 1.26 (0.165) & 9 (0.64) & $\times$ \\ 
            rpf-CV & 1.303 (0.171) & 9.441 (0.604) & $\times$ \\ 
            rf  & 1.681 (0.14) & 13.6 (1.247) & $\times$ \\ 
            sbf  & 3.972 (0.254) & 12.234 (0.688) & 1 \\ 
            gam  & 3.997 (0.251) & 12.524 (0.858) & 1 \\ 
            BART  & 1.01 (0.121) & 7.116 (0.653) & $\times$ \\ 
            BART-CV & 1.165 (0.224) & 7.897 (0.917) & $\times$ \\ 
            MARS & 3.595 (0.15) & 16.307 (1.266) & $\times$ \\ 
            1-NN & 5.839 (0.649) & 30.736 (1.933) & $\times$ \\ 
            average  & 11.471 (0.498) & 29.623 (2.046) & 0 \\ 
            \hline
		\end{tabular}
	\end{table}
	
%
%
%
	\begin{table}[ht]
		\centering
		\caption{Model 12: Pure Interaction Dense Jump Model. We report the average MSE from 100 simulations. Standard deviations are provided in brackets.}\label{lastTable} 
		\begin{tabular}{lllc}
			\hline
			Method & dim=4 & dim=10 & BI($r$) $r=0/1/2$\\ 
	        \hline
            xgboost (\texttt {depth=1})& 3.787 (0.297) & 11.106 (0.546) & 1  \\ 
            xgboost  & 1.606 (0.164) & 10.005 (0.531) & $\times$ \\ 
            xgboost-CV  & 1.768 (0.456) & 10.868 (0.648) & $\times$ \\ 
            rpf (\texttt {max\_interaction=1}) & 3.816 (0.259) & 11.246 (0.596) & 1 \\ 
            rpf (\texttt {max\_interaction=2}) & 2.005 (0.237) & 10.846 (0.488) & 2 \\ 
            rpf  & 1.441 (0.187) & 10.264 (0.433) & $\times$ \\ 
            rpf-CV  & 1.564 (0.214) & 10.582 (0.536) & $\times$ \\ 
            rf  & 1.36 (0.175) & 10.235 (0.424) & $\times$ \\ 
            sbf  & 3.928 (0.262) & 12.129 (0.639) & 1 \\ 
            gam  & 3.794 (0.271) & 11.154 (0.573) & 1 \\ 
            BART & 0.972 (0.129) & 6.835 (0.579) & $\times$ \\ 
            BART-CV  & 1.04 (0.158) & 7.208 (0.775) & $\times$ \\ 
            MARS  & 3.541 (0.289) & 11.03 (0.574) & $\times$ \\ 
            1-NN  & 4.819 (0.475) & 19.802 (1.51) & $\times$ \\ 
            average & 3.768 (0.283) & 11.03 (0.574) & 0 \\ 
            \hline
		\end{tabular}
	\end{table}

\end{document}